\documentclass[11pt]{article}

\usepackage{acl}
\usepackage{times}
\usepackage{latexsym}
\usepackage{microtype}
\usepackage{hyperref}
\usepackage{url}
\usepackage{verbatim}
\usepackage{graphicx}
\usepackage{booktabs}
\usepackage{multirow}
\usepackage{colortbl}
\usepackage{tabularx}
\usepackage{array}
\usepackage{enumitem}

\usepackage{amsmath}
\usepackage{amssymb}

\usepackage{placeins}    
\usepackage{dblfloatfix} 

\usepackage{listings}

\usepackage[dvipsnames]{xcolor}
\usepackage{pifont}
\usepackage{xcolor}
\usepackage{soul}   

\definecolor{hlsocial}{HTML}{E8DAEF}  
\definecolor{hlemotional}{HTML}{FADBD8} 
\definecolor{hlcognitive}{HTML}{D4EFDF} 
\definecolor{hlmemory}{HTML}{D6EAF8} 

\definecolor{doctorblue}{HTML}{1A5276}
\definecolor{patientgreen}{HTML}{1E8449}

\newcommand{\nsoc}[1]{{\sethlcolor{hlsocial}\hl{#1}}\textsuperscript{\tiny\textbf{[S]}}}
\newcommand{\nemo}[1]{{\sethlcolor{hlemotional}\hl{#1}}\textsuperscript{\tiny\textbf{[E]}}}
\newcommand{\ncog}[1]{{\sethlcolor{hlcognitive}\hl{#1}}\textsuperscript{\tiny\textbf{[C]}}}
\newcommand{\nmem}[1]{{\sethlcolor{hlmemory}\hl{#1}}\textsuperscript{\tiny\textbf{[M]}}}

\newcommand{\doclabel}{\textcolor{doctorblue}{\textbf{Dr:}}}
\newcommand{\patlabel}{\textcolor{patientgreen}{\textbf{Pt:}}}

\usepackage{caption}

\usepackage[most]{tcolorbox}

\definecolor{potHeader}{RGB}{225, 240, 225} 
\definecolor{potFrame}{RGB}{60, 100, 60}    
\definecolor{potBody}{RGB}{250, 252, 255}   
\definecolor{jsonKey}{RGB}{0, 50, 150}      
\definecolor{jsonString}{RGB}{0, 100, 0}    

\newtcolorbox{professionalprompt}[2][]{%
    enhanced,
    breakable,              
    colback=potBody,        
    colframe=potFrame,      
    coltitle=black,         
    colbacktitle=potHeader, 
    fonttitle=\bfseries\small\sffamily,
    title={#2},
    arc=1pt,                
    boxrule=0.6pt,          
    left=6pt, right=6pt, top=6pt, bottom=6pt, 
    toptitle=4pt, bottomtitle=4pt,
    fontupper=\footnotesize, 
    parbox=false,
    before skip=10pt,
    after skip=10pt,
    #1 
}

\lstdefinelanguage{json}{
    basicstyle=\ttfamily\scriptsize,
    columns=fullflexible,
    showstringspaces=false,
    commentstyle=\color{gray},
    keywordstyle=\color{jsonKey}\bfseries,
    stringstyle=\color{jsonString},
    breaklines=true,
    frame=none,
    morestring=[b]",
    literate=
     *{0}{{{\color{blue}0}}}{1}
      {1}{{{\color{blue}1}}}{1}
      {2}{{{\color{blue}2}}}{1}
      {3}{{{\color{blue}3}}}{1}
      {4}{{{\color{blue}4}}}{1}
      {5}{{{\color{blue}5}}}{1}
      {6}{{{\color{blue}6}}}{1}
      {7}{{{\color{blue}7}}}{1}
      {8}{{{\color{blue}8}}}{1}
      {9}{{{\color{blue}9}}}{1}
      {:}{{:}}{1}
      {,}{,}{1}
      {\{}{{{\color{black}\{}}}{1}
      {\}}{{{\color{black}\}}}}{1}
      {[}{{{\color{black}[}}}{1}
      {]}{{{\color{black}]}}}{1},
}

\newtcolorbox{QuestionnaireBox}[2][]{%
    enhanced,
    colback=potBody,
    colframe=potFrame,
    coltitle=black,
    colbacktitle=potHeader,
    fonttitle=\bfseries\small\sffamily,
    title={#2},
    arc=1pt,
    boxrule=0.6pt,
    left=6pt, right=6pt, top=6pt, bottom=6pt,
    toptitle=4pt, bottomtitle=4pt,
    fontupper=\footnotesize,
    #1
}

\newcommand{\cb}{\(\square\)} 
\newcommand{\rb}{\(\bigcirc\)} 

\newcommand{\pvar}[1]{\textcolor{NavyBlue}{\texttt{\{\{#1\}\}}}}
\newcolumntype{L}[1]{>{\raggedright\arraybackslash}p{#1}}
\newcolumntype{C}[1]{>{\centering\arraybackslash}p{#1}}

\setlist{nosep}

\title{VeriSim: A Configurable Framework for Stress-Testing Medical AI Under Patient Communication Noise}

\author{
  \textbf{Sina Mansouri}$^{*1}$ \quad
  \textbf{Mohit Marvania}$^{*1}$ \quad
  \textbf{Vibhavari Ashok Shihorkar}$^{2}$ \quad
  \textbf{Han Ngoc Tran}$^{2}$ \\[0.3em]
  \textbf{Kazhal Shafiei}$^{3}$ \quad
  \textbf{Mehrdad Fazli}$^{1}$ \quad
  \textbf{Yikuan Li}$^{2}$ \quad
  \textbf{Ziwei Zhu}$^{1}$ \\[0.6em]
  \small $^{1}$Department of Computer Science, George Mason University \\
  \small $^{2}$Department of Health Administration and Policy, College of Public Health, George Mason University \\
  \small $^{3}$Department of Science, George Mason University \\[0.4em]
  \small \texttt{\{smansou3, mmarvani, vgadhe, htran57, kshafiei, mfazli, yli94, zzhu20\}@gmu.edu}
}

\begin{document}
\maketitle
{\renewcommand\thefootnote{}\footnotetext{$^{*}$Equal contribution.}}

\begin{abstract}
Medical large language models are typically evaluated on idealized patient cases that do not reflect how real patients communicate. 
We introduce \textit{VeriSim}, a patient simulation framework that injects controllable noise along six clinically grounded communication dimensions while substantially preserving each patient's medical record.
Truth adherence is supported by a verifier that extracts atomic claims from each candidate utterance and judges them against a UMLS-grounded vector index built with BioLORD embeddings, using the retrieved atoms' structured clinical metadata (e.g., drug class, anatomical site, treats-condition relations) rather than surface-text similarity alone.
Across seven open-weight LLMs, realistic noise reduces diagnostic accuracy by 15--25 percentage points and increases conversation length by 34--55\%; the 7--8B models degrade 1.4$\times$ more than 70B+ models.
A board-certified physician and a licensed nurse rate VeriSim's conversations highly on truth, realism, clinical utility, and noise fidelity (inter-annotator agreement $\geq 0.80$ across all dimensions), and an LLM-as-judge closely tracks their ratings, enabling scalable evaluation.
We release VeriSim as an open-source framework\footnote{\url{https://github.com/mohitmarvania/VeriSim}} for rigorous stress-testing of clinical AI.
\end{abstract}


\section{Introduction}

Large language models (LLMs) are increasingly being developed for clinical applications, ranging from diagnostic support to patient communication simulation.
Recent systems have demonstrated impressive performance on static medical benchmarks, with some studies suggesting near-physician-level accuracy in controlled tasks~\cite{singhal2023large, tu2024amie, singhal2023towards}. However, while model capabilities have advanced rapidly, evaluation methodologies have remained largely static, relying on standardized, well-structured patient presentations, such as MedQA~\cite{jin2021disease}, MedMCQA~\cite{pal2022medmcqa}, and PubMedQA~\cite{jin2019pubmedqa}, that fail to reflect the ambiguity and complexity of real clinical encounters.

\begin{figure}[t]
    \centering
    \includegraphics[width=\columnwidth]{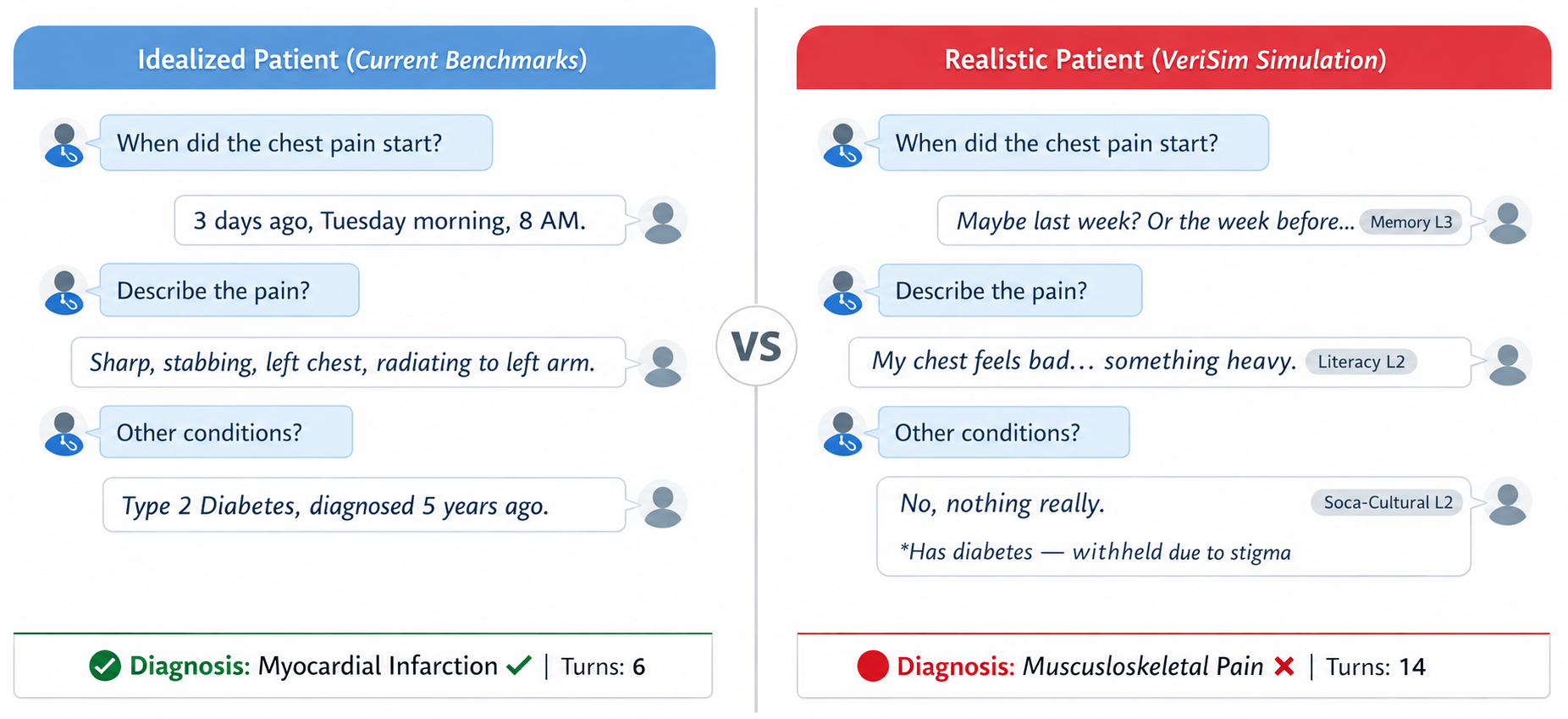}
    \caption{The Sim-to-Real gap: the \textit{same} patient case (45F, myocardial infarction) yields different outcomes depending on communication style. \textbf{Left:} idealized patient enables correct diagnosis in 6 turns. \textbf{Right:} realistic patient (VeriSim) with memory and health-literacy noise causes misdiagnosis after 14 turns. The turn counts illustrate a single case and are not corpus averages; the measured corpus-level effect is a 34--55\% increase in conversation length (Table~\ref{tab:models_results}).}
    \label{fig:sim_to_real_example}
\end{figure}

Real patients rarely communicate like textbook cases. They often forget symptom onset times, struggle to describe sensations precisely, become anxious and catastrophize, ramble about tangentially related topics, or withhold information due to stigma~\cite{street2009communication}.
For instance, a patient with myocardial infarction may report ``my chest feels heavy, maybe since last week... or was it Tuesday?'' rather than stating ``I have experienced substernal chest pain radiating to my left arm for three days.'' These communication barriers, stemming from memory limitations, health literacy gaps, emotional states, and cultural factors are not merely ``noise'' to be filtered out;
they are fundamental characteristics of authentic patient care. This discrepancy between clean benchmarks and the messy reality of practice suggests that current evaluation metrics likely overestimate the real-world robustness of medical AI systems, highlighting an urgent need for evaluation environments that bridge this ``Sim-to-Real'' gap. Figure~\ref{fig:sim_to_real_example} illustrates this gap concretely: the same myocardial infarction case is correctly diagnosed in 6 turns with an idealized patient but misdiagnosed after 14 turns when the patient exhibits realistic memory, literacy, and social-cultural noise.


\begin{figure*}[t]
    \centering
    \includegraphics[width=\textwidth]{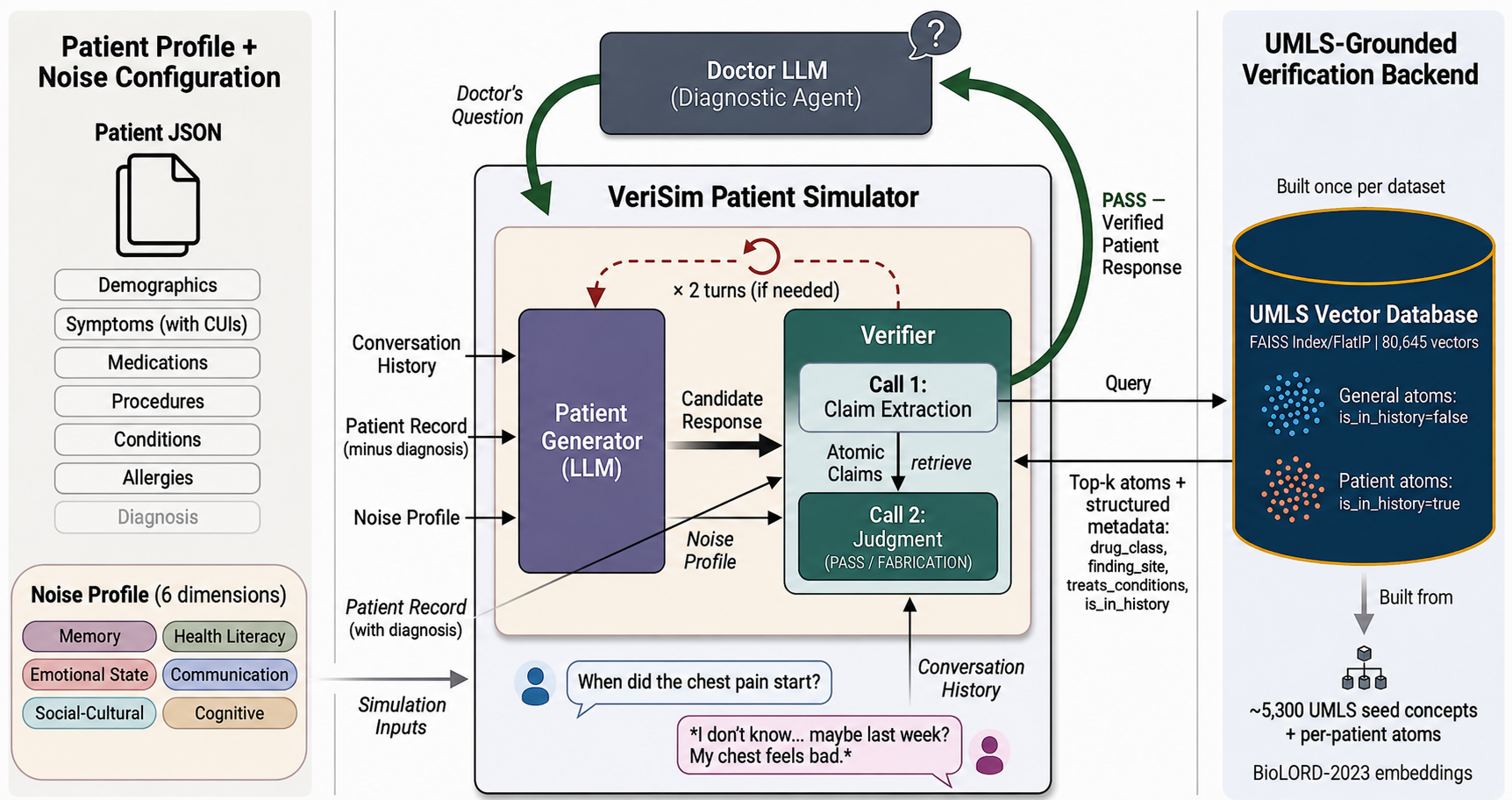}
    \caption{\textbf{VeriSim architecture.} The Patient Simulator generates noisy responses conditioned on the patient's record (minus diagnosis), conversation history, and a six-dimensional noise profile, while the Verifier enforces truth preservation through retrieval-augmented verification against a UMLS-grounded vector index. The Verifier additionally receives the diagnosis and noise profile, enabling diagnostic-leak detection and justified-denial checks. On \textsc{Fabrication}, the response is regenerated with targeted feedback (up to two attempts).}
    \label{fig:architecture}
\end{figure*}

Existing patient simulators face a trade-off between realism and factual accuracy: template-based systems~\cite{wei2018task, campillos2021virtual} are medically accurate but robotic, while prompt-driven LLM simulators~\cite{schmidgall2024agentclinic, kyung2025patientsim} are naturalistic yet cannot guarantee ground-truth adherence under controllable noise, leaving no testbed that is both behaviorally realistic and medically faithful.

We introduce VeriSim, a patient simulation framework that injects controllable communication noise — memory gaps, limited health literacy, emotional distress, and other realistic barriers — while substantially preserving each patient's medical record through a retrieval-augmented UMLS-grounded verification mechanism. By exposing controllable ``noise knobs,'' VeriSim shifts evaluation from idealized cases to diagnostic accuracy and reliability under realistic conditions.
Our contributions are threefold:
\begin{enumerate}[noitemsep]
    \item \textbf{A grounded taxonomy of patient communication noise:} We operationalize six categories of clinical barriers, memory recall, health literacy, emotional state, communication style, cognitive processing, and social-cultural factors, (Table~\ref{tab:noise_taxonomy}) each grounded in medical literature to simulate authentic clinical ambiguity.
    \item \textbf{A retrieval-augmented simulation architecture:} We propose an architecture with integrated response generation and retrieval-augmented verification that substantially decouples \textit{linguistic} expression from \textit{clinical facts}, substantially reducing hallucinations while supporting controllable noise injection.
    \item \textbf{Comprehensive evaluation revealing the Sim-to-Real gap:} We evaluate seven open-weight LLMs spanning general and medical fine-tuned variants under realistic noise conditions, demonstrating significant performance degradation (15--25 percentage-point accuracy drop) that exposes critical gaps in current AI robustness.
\end{enumerate}


\section{Related Work}

The evaluation of medical Large Language Models (LLMs) has primarily relied on static, multiple-choice datasets such as MedQA (USMLE) \cite{jin2021disease} and MedMCQA \cite{pal2022medmcqa}.
While models like Med-PaLM 2 \cite{singhal2023large} and ChatDoctor \cite{li2023chatdoctor} achieve expert-level accuracy on these benchmarks, static evaluations fail to capture the interactive dynamics of clinical diagnosis \cite{tu2024amie}.
We note that these benchmarks remain valid for assessing medical \textit{knowledge}; however, they are insufficient for evaluating clinical \textit{interaction} capabilities.
Recent efforts like SimSUM \cite{rabaey2024simsum} have attempted to link structured EHR data with unstructured notes to create richer evaluation contexts, yet they still lack the multi-turn information asymmetry inherent in real-world consultations.

\subsection{Agent-Based Patient Simulation}
To model clinical interactions, research has shifted towards agent-based simulation~\cite{wei2018task}.
Frameworks like AMIE~\cite{tu2024amie} and AgentClinic~\cite{schmidgall2024agentclinic} utilize self-play environments to optimize diagnostic dialogue.
Similarly, EHR-driven systems leverage real medical records to ground patient agents for medical education.
However, a common limitation in these systems is the ``Idealized Patient'' assumption, agents tend to provide accurate, coherent, and complete histories immediately upon inquiry.
While useful for training basic workflows, these idealized agents do not reflect the cognitive and communicative barriers that characterize challenging real-world encounters.

\subsection{Standardized Patients in Medical Education}
Our work connects to the rich tradition of Standardized Patients (SPs) in medical pedagogy, trained actors who portray patients for clinical assessments~\cite{barrows1993overview, issenberg2005features}.
VeriSim can be viewed as a ``Virtual Standardized Patient'' that provides the controllability and scalability that human SPs cannot offer, while maintaining the behavioral realism essential for valid assessment.

\subsection{Persona-Driven Simulation}
\label{sec:related_persona}
Recent work has begun to model patient variability beyond static templates. EvoPatient~\cite{du2025evopatient} and SFMSS~\cite{bao2025sfmss} evolve patient agents and simulate diverse service flows, while AIE-SAPS~\cite{liao2024automatic} uses a state-aware simulator to evaluate doctor models. Closest to our work is PatientSim~\cite{kyung2025patientsim}, which models personas along axes such as recall and cognitive confusion. These systems induce variability through prompting alone, so adherence to the record remains probabilistic: a model prompted for memory loss may invent a timeline that contradicts it. VeriSim instead enforces that adherence, checking every utterance against the patient's own record before it reaches the doctor model, which is what allows noise severity to be raised without the underlying facts drifting. Appendix~\ref{app:related_work} gives a fuller comparison.

\section{The VeriSim Framework}

\subsection{Taxonomy of Clinical Noise}
\label{sec:taxonomy}

To bridge the simulation-to-reality gap, we operationalize the complexity of patient interaction into six distinct, independently controllable dimensions (pillars).
Unlike generic noise injection, our taxonomy is grounded in medical communication literature, ensuring that the simulated barriers reflect authentic clinical phenomena. Table~\ref{tab:noise_taxonomy} details the theoretical grounding and behavioral manifestation of each pillar.
The individual phenomena are themselves well established in clinical communication research; what we contribute is their operationalisation. We select six barriers arising from distinct underlying mechanisms and render each as an independently tunable severity axis, so that an encounter can be specified, reproduced, and ablated one barrier at a time, which to our knowledge no prior patient simulator supports.
We treat these pillars as \textit{orthogonal axes}, enabling the generation of diverse patient profiles.
While some overlap may exist (e.g., Health Literacy affects vocabulary while Cognitive Processing affects reasoning/beliefs), we define clear boundaries: Literacy governs \textit{vocabulary and conceptual understanding}, while Cognitive Processing governs \textit{reasoning patterns and belief formation} (e.g., confirmation bias from internet research).
We define five severity levels (0--4) for each pillar, mapping qualitative clinical descriptions to quantitative parameters.
\begin{table*}[t]
\centering
\footnotesize
\begin{tabular}{@{}l p{5.4cm} p{6.0cm}@{}}
\toprule
\textbf{Noise Pillar} & \textbf{Clinical Basis \& Evidence} & \textbf{Simulated Behavior (Example)} \\ 
\midrule
\textbf{1.
Memory \& Recall} & Recall of one's own history is unstable: patients underreport prior symptoms, merge separate episodes, and recall events that did not occur~\cite{barsky2002forgetting}. Stroke recall sensitivity is 17.4\% against neuroimaging~\cite{day2020assessing}; retention of clinician-provided information is similarly poor~\cite{kessels2003patients}. & \textit{``It started... maybe last week?
Or two weeks ago?''} \\
\addlinespace
\textbf{2. Health Literacy} & 36\% of US adults possess limited health literacy \cite{kutner2006health}.
& Uses ``sugar'' for diabetes; ``stomach hurting'' for epigastric pain. \\
\addlinespace
\textbf{3.
Emotional State} & Anxiety amplifies symptom perception through somatosensory amplification~\cite{barsky1990amplification, craig2009social}.
& \textit{``It hurts so much, I'm sure it's a heart attack!''} \\
\addlinespace
\textbf{4.
Communication Style} & Standard interviewing references observe that patients rarely present symptoms in the organized, logical form found in medical texts, and that some require the physician to set limits on a rambling history~\cite{lichstein1990interview}.
& Rambles about unrelated topics when asked direct questions. \\
\addlinespace
\textbf{5.
Social-Cultural} & Stigma leads to non-disclosure of sensitive health behaviors \cite{kulesza2013review}. & Denies alcohol use initially; admits after empathetic probing.
\\
\addlinespace
\textbf{6. Cognitive Processing} & A substantial proportion of patients engage in online self-diagnosis prior to clinical consultation, often leading to anxiety amplification \cite{starcevic2013cyberchondria}.
& \textit{``I read online this is definitely Lupus.''} \\ 
\bottomrule
\end{tabular}
\caption{The six pillars of patient communication noise, each grounded in clinical evidence and independently controllable.}
\label{tab:noise_taxonomy}
\end{table*}

Detailed parameter mappings for each severity level and implementation prompts are provided in Appendix~\ref{app:noise_params}.
\subsection{System Architecture}
\label{sec:architecture}

We propose a unified patient simulation framework that substantially decouples \textit{linguistic expression} from \textit{clinical facts}.
We use ``decouples'' in a measurable sense: as severity rises, the wording of an in-record fact should diverge while its clinical content stays faithful. Raising both assigned pillars from $L_0$ to $L_3$ moves surface variation from 0.12 to 0.63 while ground-truth preservation falls only from 99.6\% to 93.3\% (Appendix~\ref{app:decoupling}), so the wording changes steeply while the facts hold.
As illustrated in Figure~\ref{fig:architecture}, the system appears externally as a single conversational agent but internally operates via a three-step \textit{Generate-Verify-Refine} loop.
This design ensures that increased behavioral realism does not compromise medical validity, a critical limitation of prompt-only simulation approaches~\cite{ji2023survey}.


The three steps are: (1) the \textit{Generator} produces a noisy patient response conditioned on the noise profile; (2) the \textit{Verifier} checks whether the response is medically valid by retrieving evidence from a UMLS-grounded vector database and reasoning over it with an LLM; and (3) if verification fails, the system \textit{Refines} by regenerating with targeted feedback.
We additionally enforce \textit{Information Asymmetry}: the generator is blinded to the diagnosis, mirroring real patients who experience symptoms without knowing their condition.

\subsubsection{Step 1: Noisy Response Generation}
\label{sec:generation}

Given a doctor's question, the generator produces a candidate response $\hat{r}$ conditioned on three inputs: the conversation history $\mathcal{H}$, the patient's medical record $\mathcal{E}$ (explicitly \textit{excluding} the final diagnosis to prevent data leakage), and the noise profile $\mathcal{N}$.
Conditioning on the full history makes the patient stateful rather than one-shot: behaviour accumulates across turns, so a patient under Emotional-State noise becomes progressively more distressed as the encounter proceeds (Appendix~\ref{app:conversations}).

The noise profile acts as a \textit{composite style constraint} that instructs the model to manifest specific behavioral deficits during generation.
Under $\mathcal{N} = \{\text{Memory Level 3}, \text{Health Literacy Level 2}\}$, for instance, a clean answer of ``three days ago, Tuesday morning'' becomes ``I don't know... maybe last week? My chest feels bad'', exhibiting both a confused timeline and simplified vocabulary.
The challenge is that such responses may inadvertently introduce fabricated symptoms: a model instructed to ``be confused'' might invent one (``my leg hurts too'') that contradicts the ground truth.
This motivates the verification step.

\subsubsection{Step 2: Retrieval-Augmented Verification}
\label{sec:verification}

The verifier must distinguish \textit{controllable noise}, which should be permitted, from \textit{medical fabrication}, which must be blocked.
For a patient whose record contains chest pain but no diabetes, ``my chest feels heavy'' and ``I take my blood pressure pill'' are valid lay renderings of in-record facts, while ``I take my diabetes pill'' invents a medication.
A naive string matcher would reject the first two; a fully unconstrained LLM verifier would accept the third.
We address this through retrieval-augmented verification grounded in the Unified Medical Language System (UMLS)~\cite{bodenreider2004umls}, the biomedical knowledge graph maintained by the U.S. National Library of Medicine that integrates over 200 source vocabularies (SNOMED CT, RxNorm, ICD-10-CM, CPT).
The verifier operates in two phases (Figure~\ref{fig:architecture}): an \textit{offline} phase builds a vector index over UMLS atoms; a \textit{runtime} phase queries this index to judge each candidate response.

\paragraph{Phase A: Offline Vector Database Construction.}
The vector index is the verifier's grounding source, and it holds two complementary sets of UMLS atoms distinguished by an \texttt{is\_in\_history} flag.
The general subset (\texttt{is\_in\_history}=false) covers clinically common symptoms, medications, procedures, and conditions, supplying the vocabulary against which any utterance can be grounded; the patient subset (\texttt{is\_in\_history}=true) expands every entry of a given patient's record into its full set of atoms, including synonyms and lay variants, each tagged with the originating \texttt{patient\_id}.
Every atom additionally carries structured clinical metadata extracted offline from SNOMED CT and RxNorm relations, such as \texttt{drug\_class} and \texttt{treats\_conditions} for a medication or \texttt{finding\_site} for a symptom, which gives the verifier evidence beyond surface-text similarity.
Atoms are embedded with BioLORD-2023~\cite{remy2024biolord}, a sentence-transformer trained on UMLS concept-definition pairs that outperforms general biomedical encoders on patient-language normalization, and indexed with FAISS~\cite{johnson2019billion} for exact cosine retrieval.
Appendix~\ref{app:umls_details} gives the construction procedure in full.

\paragraph{Phase B: Runtime Verification.}
Given a candidate response $\hat{r}$, the verifier runs two LLM calls bridged by a retrieval step.

\textit{Call 1 (Claim extraction).}
The LLM decomposes $\hat{r}$ into a list of atomic medical claims $\{c_1, \ldots, c_n\}$, where each $c_i$ is one self-contained assertion (e.g., \textit{``patient takes a diabetes medication''}).
Non-medical content (emotional expressions, hesitation, filler) is discarded.

\textit{Retrieval.}
For each claim $c_i$, we embed $c_i$ with BioLORD and retrieve the top $k{=}5$ nearest atoms from the FAISS index, along with their full metadata (\texttt{is\_in\_history}, \texttt{patient\_id}, and category-specific clinical attributes such as \texttt{drug\_class} and \texttt{treats\_conditions}).

\textit{Call 2 (Judgment).}
The LLM is now given everything it needs to judge each claim: the original utterance, the extracted claims, the top-$k$ retrieved atoms with metadata for each claim, and the patient's full ground-truth record.
It issues a per-claim verdict in \{\textsc{Pass}, \textsc{Fabrication}\}, and the overall response is marked \textsc{Regenerate} if any claim is a fabrication.
A claim passes when a top-retrieved atom is in-history for this patient, or when the retrieved atoms' clinical attributes overlap the record, for instance when a retrieved drug's class matches one the patient takes.
Underspecified claims whose retrievals all fall below $\tau{=}0.55$ (e.g., \textit{``I feel weird''}) also pass, since they assert nothing verifiable.
Only when neither condition holds is the claim a fabrication.

This structured-metadata design is a deliberate departure from standard RAG verification, which typically gives the LLM only retrieved text passages.
Exposing the verifier to clinical attributes lets it confirm a lay expression when its top retrieval is not a perfect lexical match (via attribute overlap), and detect fabrications even when surface text looks plausible (via attribute mismatch).

The verifier additionally enforces three global constraints: (1) \textit{justified denial} (symptom omissions must match the noise profile), (2) \textit{demographic invariance} (age, sex, occupation remain stable), and (3) \textit{history consistency} (no contradictions with prior turns).
Full prompt templates are provided in Appendix~\ref{app:prompts}.

On \textsc{Regenerate}, the generator is re-invoked with targeted feedback (up to $k{=}2$ attempts); across a held-out subset, no fabrication-flagged claim reached the Doctor LLM (Appendix~\ref{app:refinement}). The full three-step pipeline therefore satisfies both the noise profile (behavioral realism) and the UMLS-grounded constraints (medical validity).



\subsection{Controllability and Configuration}
\label{sec:controllability}

Drawing on principles from controllable text generation~\cite{keskar2019ctrl}, we formalize the simulation as a configurable system with explicit, interpretable parameters.
Complete mappings are in Appendix~\ref{app:noise_params}.


\paragraph{Experimental Protocol.}
Each patient receives \textit{two noise types} with severity levels drawn from $L_1$--$L_3$.
Levels $L_0$ (ideal) and $L_4$ (extreme) are reserved for ablation studies and qualitative stress-test examples.
The choice of two simultaneously active pillars is empirical rather than arbitrary: a single barrier already costs 9.7 accuracy points but understates the difficulty of a realistic encounter, while three compress into one short patient turn and begin to cost realism for only a modest further accuracy drop (Appendix~\ref{app:pillar_ablation}).
Configurations are serialized as JSON with fixed random seeds for reproducibility.
\section{Experimental Setup}

\subsection{Datasets}
\label{sec:data}

We constructed a diverse evaluation set of 300 patient cases by extracting and harmonizing records from two complementary sources: DDXPlus \cite{fansi2022ddxplus}, a synthetic differential diagnosis dataset, and MIMIC-IV-ED \cite{johnson2023mimicived}, a real-world emergency department database.
This sample size is consistent with established benchmarks for agentic clinical simulation, exceeding the 149 cases used in AMIE's primary physician comparison~\cite{tu2024amie} and comparable to AgentClinic's 260-case evaluation~\cite{schmidgall2024agentclinic}.
For each case, we extracted demographics (age, sex), ground-truth diagnosis with ICD-10 code, and presenting symptoms.
All cases were converted to a unified JSON schema (~\ref{fig:patient_example}) compatible with our simulator input format (Hyperparameters ~\ref{tab:hyperparams}).
This source-agnostic representation ensures our framework generalizes across both synthetic and real clinical data.
\paragraph{UMLS Vector Database Construction.}
As part of dataset preparation, we construct the UMLS-grounded vector index described in Section~\ref{sec:verification}.
For the general medical subset, we resolve a curated seed list of $\sim$5,300 clinically relevant concepts spanning symptoms (SNOMED CT semantic types T184, T033), medications (RxNorm), procedures (CPT, SNOMED), and conditions (ICD-10-CM, SNOMED) to UMLS CUIs via the UMLS REST API.
For each patient case, we additionally retrieve all atoms associated with the patient's ground-truth symptoms, medications, procedures, conditions, and allergies, tagging each with the originating \texttt{patient\_id} and \texttt{is\_in\_history}=true.
Across both subsets, we extract every atom's surface text together with its structured clinical relations from SNOMED CT and RxNorm (e.g., \texttt{finding\_site}, \texttt{has\_ingredient}, \texttt{drug\_class}), then embed all unique atom texts with BioLORD-2023 and index them with FAISS \texttt{IndexFlatIP}.
The resulting index contains 80,645 unique vectors (768-dimensional) over 100,135 atoms, occupies 236MB, and is built once per dataset.
At runtime, retrieval over this index takes milliseconds per query and adds no UMLS API dependency.
Our evaluation encompasses 300 patient cases $\times$ 7 doctor models = \textit{2,100 diagnostic conversations}.

\subsection{Models}
\label{sec:models}

\paragraph{Patient Simulator.}
The unified Patient Simulator (Generator + Verifier) uses Llama-3.1-70B-Instruct to ensure sufficient reasoning capability for retrieval-augmented verification (prompts in Appendix~\ref{app:prompts}, Figures~\ref{fig:patient_simulator_prompt}--\ref{fig:semantic_prompt}).
The same model handles both response generation and UMLS-grounded truth verification. While the same base model is used, the verification task is fundamentally distinct: the verifier receives the retrieved UMLS atoms with their structured clinical metadata and the patient's full ground-truth record (symptoms, medications, procedures, conditions, allergies), all of which are withheld from the generator. This transforms the model into a grounded fact-checker rather than a self-assessor, and Section~\ref{sec:ablation} evaluates it directly against human labels. 


\subsection{Metrics}
\label{sec:metrics}

\paragraph{Diagnostic Performance.}
We measure \textit{Top-1 Accuracy}, defined as a match between the doctor model's final diagnosis and the ground-truth diagnosis.
Doctor models emit free text (``Final Diagnosis: [condition]''; Figure~\ref{fig:doctor_llm_prompt}), so the match is resolved by a three-stage cascade: the prediction is normalised and mapped through UMLS to an ICD-10 code compared at category level, which credits ``heart attack'' and ``MI'' as one answer; failing that, prediction and reference match if they resolve to a shared CUI; residual borderline cases go to an LLM judge. Against 100 human-labelled pairs the cascade reproduces the human decision 96\% of the time ($\kappa = 0.91$), and the clean-to-noisy gap is stable across matchers of differing strictness (Appendix~\ref{app:matcher}).
This metric is consistent with prior work in diagnostic dialogue evaluation~\cite{tu2024amie, schmidgall2024agentclinic} and reflects the clinical requirement for precise diagnostic conclusions.

\paragraph{Conversation Efficiency.}
We track two complementary metrics: (1) \textit{Average Turns}, the total number of conversational exchanges before the doctor model issues a final diagnosis, which measures how efficiently the model gathers diagnostic information under noisy conditions~\cite{tu2024amie}; and (2) \textit{Average Turns Increase} ($\Delta$Turns), the relative increase in conversation length between clean and noisy conditions, which quantifies the additional diagnostic burden imposed by patient communication barriers.

\subsection{Evaluation Protocol}
\label{sec:eval_protocol}

\paragraph{Human Evaluation.}
One board-certified physician and one licensed nurse independently evaluated a stratified 300 patient conversations.
Both evaluators possess formal training in health informatics and have substantial real-world clinical experience.
Evaluators received conversation transcripts, ground-truth symptoms, the assigned noise profile, and a detailed rubric with specific anchor examples for each score level~\ref{fig:questionnaire}.
This structured approach reduced subjective interpretation, enabling objective assessment of whether patient behavior matched the assigned noise configuration.
\paragraph{Sample Size Justification.}
Power analysis indicates that for the observed effect sizes (Cohen's $d > 1.2$ for truth preservation, $d > 0.8$ for realism), 300 samples provide statistical power $> 0.99$ to detect significant differences at $\alpha = 0.05$.
Combined with the high inter-annotator agreement, this sample size is sufficient to validate the LLM-as-a-Judge protocol.
\paragraph{LLM-as-Judge.}
We employ Claude Opus-4.5 as an automated evaluator using the same questionnaire~\ref{fig:questionnaire} as human evaluators.
The judge receives full context including noise profiles and uses Chain-of-Thought prompting~\ref{fig:llm_as_judge_prompt} to generate reasoning before scoring \cite{liu2023geval}.
We validated this approach by measuring agreement with human evaluators.
\section{Results and Analysis}

\subsection{Diagnostic Performance Under Noisy Patients}
\label{sec:main_results}
Table~\ref{tab:models_results} presents diagnostic performance across all models under clean (Level 0) and noisy (Levels 1--3) patient conditions.

\begin{table*}[t]
\centering
\small
\begin{tabular}{@{}lcccccc@{}}
\toprule
\textbf{Model} & \textbf{Size} & \textbf{Type} & \textbf{Clean} & \textbf{Noisy} & \textbf{$\Delta$Acc} & \textbf{$\Delta$Turns} \\
\midrule
Qwen-2.5-72B~\cite{yang2024qwen25} & 72B & General & 84.5 & 69.2 & -15.3$^{***}$ & +34.5\% \\
Llama-3.1-70B~\cite{grattafiori2024llama} & 70B & General & 82.1 & 65.5 & -16.6$^{***}$ & +35.3\% \\
Meditron-70B~\cite{chen2023meditron} & 70B & Medical & 78.4 & 62.8 & -15.6$^{***}$ & +35.2\% \\
OpenBioLLM-70B~\cite{saama2024openbiollm} & 70B & Medical & 79.2 & 63.1 & -16.1$^{***}$ & +36.0\% \\
\midrule
Llama-3.1-8B~\cite{grattafiori2024llama} & 8B & General & 61.8 & 40.2 & -21.6$^{***}$ & +48.8\% \\
BioMistral-7B~\cite{labrak2024biomistral} & 7B & Medical & 64.2 & 41.8 & -22.4$^{***}$ 
& +50.0\% \\
Mistral-7B~\cite{jiang2023mistral} & 7B & General & 58.0 & 33.5 & -24.5$^{***}$ & +55.1\% \\
\bottomrule
\end{tabular}
\caption{Doctor LLMs evaluated and their diagnostic performance under clean (Level 0) vs. noisy (Levels 1--3) patient conditions.
Clean/Noisy columns report Top-1 accuracy (\%). All differences are statistically significant (paired t-test, $^{***}p<0.001$).}
\label{tab:models_results}
\end{table*}

\paragraph{All Models Degrade Under Noise.}
Without exception, every model shows significant performance degradation when interacting with noisy patients.
Even the strongest model (Qwen-2.5-72B) experiences a 15.3 percentage-point drop in diagnostic accuracy, while conversation length increases by 34.5\%.
\paragraph{Medical Fine-tuning and Noise Robustness.}
Medical fine-tuning on clean, textbook-style clinical text does not transfer to noisy patient interactions: BioMistral-7B ($-22.4$\%) shows comparable degradation to Mistral-7B ($-24.5$\%), and 70B medical variants (Meditron, OpenBioLLM) excel at knowledge retrieval but were not exposed to patients exhibiting memory gaps or emotional distress during training. This suggests existing medical corpora lack the communicative diversity of real clinical encounters, a gap VeriSim itself could help close by generating noisy training data.

\subsection{Simulation Quality Validation}
\label{sec:judge_validation}

We validate the quality of VeriSim's simulated patient interactions through both human expert evaluation and automated LLM-based assessment.

\paragraph{High-Quality Simulation.}
Human evaluators rated our simulated conversations highly across all four evaluation dimensions.
Truth preservation scored highest, with both evaluators confirming that 93.3\% of evaluated conversations contained no hallucinated symptoms or ground-truth violations.
Realism assessments averaged 4.05/5.0, indicating that evaluators found the simulated patients were behaviorally convincing despite the UMLS-grounded constraints.
Clinical utility scores confirmed that the conversations are appropriately challenging for diagnostic training.
Noise fidelity is worth reporting as scores and not only as agreement, since high agreement on \textit{low} scores would mean raters concurred that the injected noise was rendered poorly. The per-pillar means are instead consistently high (3.6--4.3; overall 4.10 and 4.05 for the two evaluators), with Communication Style rated lowest, consistent with tangential responding being the most subjective barrier to assess (Appendix~\ref{app:noise_fidelity}).

\paragraph{LLM-as-Judge Reliability.}
To enable scalable evaluation beyond the 300 human-annotated conversations, we validated an LLM-based judge (Claude Opus-4.5) against human evaluators.
Table~\ref{tab:agreement} presents agreement across all dimensions.

\begin{table}[t]
\centering
\small
\begin{tabular}{@{}lccc@{}}
\toprule
\textbf{Dimension} & \textbf{H1-H2} & \textbf{LLM-H} & \textbf{Metric} \\
\midrule
Truth Preservation & 0.91 & 0.87 & $\kappa$ \\
Realism Assessment & 0.84 & 0.78 & $r$ \\
Clinical Utility & 0.80 & 0.74 & $r$ \\
Noise Fidelity & 0.83 & 0.77 & $\kappa$ \\
\bottomrule
\end{tabular}
\caption{Inter-annotator agreement between human evaluators (H1-H2) and LLM-Human agreement (LLM-H, averaged over both evaluators).
$\kappa$ = Cohen's Kappa; $r$ = Pearson correlation.}
\label{tab:agreement}
\end{table}

LLM-Human agreement approaches Human-Human agreement across all dimensions, with a consistent gap of only 0.05--0.06 points.
For truth preservation, the most objective dimension LLM-Human agreement reaches $\kappa = 0.87$ (``almost perfect'' on the Landis \& Koch scale), indicating that the automated judge reliably identifies hallucinations and ground-truth violations.
More subjective dimensions such as Noise Fidelity show expected moderate-to-strong agreement ($\kappa = 0.77$), consistent with typical human-human agreement ranges in medical evaluation~\cite{mchugh2012interrater}.

\subsection{Ablation: Verifier Effectiveness}
\label{sec:ablation}

To validate the contribution of our retrieval-augmented UMLS verification module, we conduct an ablation study comparing three system configurations (Table~\ref{tab:ablation}).
Hallucination rates were assessed across all 2,100 conversations using our validated LLM-as-Judge protocol, with human annotators independently confirming reliability on 300 conversations ($\kappa = 0.87$, Table~\ref{tab:agreement}).
\begin{table}[t]
\centering
\small
\begin{tabular}{@{}lcccc@{}}
\toprule
\textbf{Metric} & \textbf{No Ctrl} & \textbf{Prompt} & \textbf{Ours} \\
\midrule
Halluc. Rate $\downarrow$ & 24.2\% & 17.8\% & \textbf{6.7\%} \\
Consist. Rate $\uparrow$ & 67.5\% & 79.8\% & \textbf{93.3\%} \\
Realism (1-5) $\uparrow$ & 4.18 & 4.11 & 4.05 \\
\bottomrule
\end{tabular}
\caption{Ablation study on Verifier configurations.
\textit{No Ctrl} = unconstrained generation; \textit{Prompt} = prompt-based verification without UMLS; \textit{Ours} = full retrieval-augmented UMLS verification.
Consistency Rate measures the absence of hallucinations, contradictions, demographic drift, and history inconsistency (i.e., it is stricter than the complement of hallucination rate).
Our approach achieves 72.3\% relative reduction in hallucination rate with only 3.1\% realism trade-off.}
\label{tab:ablation}
\end{table}

\paragraph{No Controller.}
Without verification, the Patient Generator hallucinates symptoms in 24.2\% of responses fabricating symptoms outside the ground-truth diagnosis.
This rate aligns with reported hallucination rates (16--31\%) for unconstrained LLMs in medical domains~\cite{ji2023survey}.
\paragraph{Prompt-Only Controller.}
Prompt-based verification reduces hallucinations to 17.8\%, but remains insufficient, permitting semantically plausible symptoms that fall outside the patient's actual condition.
\paragraph{Retrieval-Augmented Verification (Ours).}
Our full system achieves 6.7\% hallucination rate, a \textit{3.6$\times$ reduction} compared to no controller.
The 93.3\% consistency rate demonstrates reliable adherence to ground truth across conversations.

\paragraph{Verifier Independence.}
Generator and verifier share a base model (Section~\ref{sec:models}), so the residual 6.7\% could reflect a model agreeing with itself. Two results argue otherwise. Against expert labels on a 200-claim benchmark ($\sim$40\% fabrications), the verifier reaches precision 0.93, recall 0.90, $F_1$ 0.91 ($\kappa = 0.87$ on claims; Appendix~\ref{app:verifier_eval}); recall is the operative quantity, since missed fabrications are what would inflate the 6.7\%. Swapping both roles to Qwen-2.5-72B leaves the pipeline near-identical (6.7\%$\rightarrow$7.1\% hallucination, 93.3\%$\rightarrow$92.6\% consistency, 4.05$\rightarrow$4.08 realism), so truth preservation follows from the architecture, not one model family.

\paragraph{Safety-Realism Trade-off.}
The modest realism decrease (4.18 $\rightarrow$ 4.05, -3.1\%) represents an acceptable cost for 72.3\% hallucination reduction.
The 0.09-point difference falls within the variation we observe between our two raters, whereas the hallucination reduction is a 3.6$\times$ change.
In medical evaluation, \textit{functional fidelity} (accurate symptom presentation) takes precedence over \textit{linguistic fluidity}.



\subsection{Key Findings and Implications}
\label{sec:implications}

\paragraph{The Idealized-to-Noisy Gap.}
The 15--25 percentage-point accuracy drop observed across all models (Table~\ref{tab:models_results}) suggests that current evaluation protocols may overestimate clinical capability when conditions deviate from idealized benchmarks. This degradation reflects a combination of factors: literature-grounded communication ambiguity, deliberate information withholding (e.g., under Social-Cultural noise), and the addition of plausible but non-diagnostic content (under Emotional and Communication Style noise). A side-by-side example of how the same ground-truth symptoms surface differently under clean versus noisy conditions is provided in Appendix~\ref{app:clean_vs_noisy}

\paragraph{Scale and Deployment.}
Even the best 70B+ models lose roughly one-sixth of their diagnostic accuracy under realistic noise, and smaller 7B--8B models collapse much further; scaling alone is insufficient and raises concerns about deploying small models in patient-facing applications. Beyond evaluation, VeriSim can also generate noisy dialogues for curriculum-learning strategies that train medical LLMs robust to such conditions.

\section{Conclusion}
We introduced VeriSim, a patient simulation framework pairing a six-pillar noise taxonomy with retrieval-augmented UMLS verification, enabling controllable stress-testing of medical AI while substantially preserving medical validity. Experiments across seven open-weight LLMs reveal a substantial idealized-to-noisy gap that current benchmarks fail to capture. Future work includes fine-tuning the retrieval encoder on patient-language corpora, extending to multilingual settings, and adding dimensions the current taxonomy does not expose, such as patient-initiated questioning and severity that shifts within a conversation as rapport develops.

\section*{Limitations}

Our error analysis of diagnostic failures reveals three primary failure modes.
The most prevalent is \textit{noise-induced misdirection} (43.6\%), where tangential patient responses, particularly under Communication Style and Emotional State noise, lead doctor models down incorrect differential diagnosis paths.
For example, a patient with pneumonia who rambles about a recent family trip may cause the model to pursue travel-related infections rather than community-acquired pneumonia.
The second mode is \textit{semantic ambiguity} (38.2\%), where colloquial terms (e.g., ``sugar problems'' instead of ``Type 2 Diabetes'') are not resolved through targeted clarification questions.
The remaining failures (18.2\%) involve \textit{premature diagnostic commitment}, where models reach a diagnosis before gathering sufficient discriminating information, particularly under Social-Cultural noise where patients withhold key symptoms.

\paragraph{Coverage of Modern Colloquial Language.}
Our vector index relies on BioLORD-2023 embeddings, which were trained on UMLS concept-definition pairs and demonstrate strong performance on standard patient-language benchmarks (e.g., CADEC, PsyTAR)~\cite{remy2024biolord}.
However, BioLORD's training corpus is not designed to cover the most recent internet-era slang and informal expressions (e.g., social-media abbreviations, generation-specific idioms).
While our evaluation focuses on acute physiological presentations where standard colloquial terminology (``chest pressure,'' ``stomach ache,'' ``heart cath thing'') is well-covered, fully capturing the long tail of contemporary patient slang may require fine-tuning the embedding model on social-media or patient-forum corpora.

\paragraph{Human Evaluation Scale.}
Our human evaluation involved two clinical domain experts assessing 300 conversations.
While the inter-annotator agreement (Cohen's $\kappa \geq 0.83$ for categorical dimensions; Pearson $r \geq 0.80$ for ordinal dimensions) demonstrates reliable assessment consistent with typical medical evaluation benchmarks, expanding to additional annotators and larger samples would strengthen generalizability.
The two-expert design reflects the high cost of clinical expertise in medical NLP evaluation.

\paragraph{Retrieval-Augmented Verifier Trade-offs.}
Our retrieval-augmented verification approach delegates final judgment to an LLM that reasons over retrieved UMLS atoms and their structured clinical metadata.
While the structured metadata reduces sensitivity to retrieval rank, the LLM verifier may still exhibit inconsistent judgments on edge cases where multiple retrieved atoms have overlapping but non-identical clinical attributes.
Empirically the system is effective (6.7\% hallucination rate), but future work could explore ensemble verification, calibrated confidence thresholds, or structured rule-based fallbacks to further improve reliability.
Additionally, our evaluation is limited to English-language interactions and may not generalize to multilingual clinical settings, although BioLORD-2023 has a multilingual variant~\cite{remy2024biolord} that could be adapted in future work.

\section*{Ethical Considerations}

This research uses synthetic patient data (DDXPlus) and de-identified clinical records (MIMIC-IV-ED) in compliance with data use agreements.
No live patient interactions were conducted, and no identifiable patient data were used. Human evaluators provided informed consent.
Our framework is intended for research evaluation purposes and should not be used as a substitute for clinical judgment or patient care.

\section*{AI Assistance}
AI tools were used for editorial refinement and programming assistance. All technical contributions and findings were verified by the authors.

\section*{Acknowledgments}
\label{sec:ack}

This work is in part supported by NSF grant IIS-2452129. We acknowledge the support by resources provided by the Office of Research Computing at George Mason University (\url{https://orc.gmu.edu}) and funded in part by grants from the NSF (Award Number 2018631). Any opinions, findings, and conclusions or recommendations expressed in this material are solely those of the authors.

\bibliography{references}
\bibliographystyle{acl_natbib}

\clearpage
\appendix

\setcounter{table}{0}
\renewcommand{\thetable}{A\arabic{table}}
\setcounter{figure}{0}
\renewcommand{\thefigure}{A\arabic{figure}}

\section{Prompt Templates}
\label{app:prompts}

We employ a structured prompting strategy across all components.
Variables injected at runtime are denoted by \pvar{variable\_name}.

\begin{figure}[h]
    \begin{professionalprompt}{System Instruction: Patient Generator}
    \textbf{[ROLE]}
    You are a simulated patient in a clinical encounter.
    
    \textbf{[PROFILE]}
    \textbf{Demographics:} You are a \pvar{age}-year-old \pvar{sex}.\\
    \textbf{Symptoms:} \pvar{patient\_symptoms} \\
    \textbf{Current medications:} \pvar{patient\_medications} \\
    \textbf{Past medical conditions:} \pvar{patient\_conditions} \\
    \textbf{Past procedures:} \pvar{patient\_procedures} \\
    \textbf{Allergies:} \pvar{patient\_allergies} \\
    \textbf{Knowledge Constraint:} You do NOT know your diagnosis. If asked directly, express uncertainty. Never use the term ``\pvar{diagnosis\_label}'' or any clinical synonym.
    
    \textbf{[NOISE PROFILE]}
    Apply the following communication barriers naturally:
    \begin{itemize}[leftmargin=15pt, label=$\blacktriangleright$]
        \item \textbf{\pvar{noise\_type\_1} (Level \pvar{level\_1}):} \pvar{behavior\_1}
        \item \textbf{\pvar{noise\_type\_2} (Level \pvar{level\_2}):} \pvar{behavior\_2}
    \end{itemize}

    \textbf{[CORE INSTRUCTIONS]}
    \begin{enumerate}[leftmargin=15pt]
        \item \textbf{Stay in character:} Respond naturally as a real patient would.
        \item \textbf{Truth preservation:} Only mention symptoms, medications, conditions, procedures, or allergies that are in your profile above. Never fabricate medical facts.
        \item \textbf{Information Asymmetry:} Do not volunteer symptoms unprompted. Answer only specific questions asked.
        \item \textbf{Consistency:} Maintain history consistency throughout the conversation.
        \item \textbf{No ``AI'' Language:} Never reference being an AI, the simulation, or the noise profile.
    \end{enumerate}

    \vspace{2pt}
    \textbf{[TASK]}
    Generate the patient's next response based on the conversation history. Keep it under 50 words.
    \end{professionalprompt}
    \caption{Patient Generator Prompt Template}
    \label{fig:patient_simulator_prompt}
\end{figure}

\begin{figure}[h]
    \begin{professionalprompt}{System Instruction: Verifier Call 1 -- Claim Extraction}
    \textbf{[ROLE]}
    You are a medical claim extractor. Given a doctor's question and the patient's response, your task is to decompose the patient's response into atomic medical claims that can be independently verified.

    \textbf{[INPUT]}
    \textbf{Doctor's question:} \pvar{doctor\_question} \\
    \textbf{Patient's response:} \pvar{candidate\_response}

    \textbf{[DEFINITION OF AN ATOMIC CLAIM]}
    An atomic claim is a single, self-contained medical assertion. Examples:
    \begin{itemize}[leftmargin=15pt, label=$\blacktriangleright$, noitemsep]
        \item ``patient has chest tightness''
        \item ``patient takes a beta blocker''
        \item ``patient had a coronary stent placed''
        \item ``patient is allergic to penicillin''
    \end{itemize}
    \textbf{Discard} non-medical content:
    \begin{itemize}[leftmargin=15pt, label=$\circ$, noitemsep]
        \item Emotional expressions (``I'm so scared'', ``please help me'')
        \item Conversational filler (``well, uh, you know'')
        \item Hedging without medical content (``maybe'', ``I think'')
        \item Questions back to the doctor
    \end{itemize}

    \textbf{[INSTRUCTIONS]}
    \begin{enumerate}[leftmargin=15pt, noitemsep]
        \item Identify every medically meaningful assertion in the response.
        \item Express each as a single, self-contained statement in third person.
        \item Preserve colloquial phrasing where the patient used it (``blood pressure pill'' rather than rewriting to ``antihypertensive medication'').
        \item If the response contains no medical claims, return an empty list.
    \end{enumerate}

    \textbf{[OUTPUT FORMAT]}
    Return a JSON list of strings, one per claim. Output \textbf{only} the JSON, with no preamble.
    \begin{lstlisting}[language=json]
["claim 1", "claim 2", ...]
    \end{lstlisting}
    \end{professionalprompt}
    \caption{Verifier Call 1: atomic claim extraction prompt template.}
    \label{fig:verifier_extract_prompt}
\end{figure}

\begin{figure*}[t]
    \begin{professionalprompt}{System Instruction: Verifier Call 2 -- Judgment}
    \textbf{[ROLE]}
    You are a medical verifier. Your task is to decide whether a patient's response is consistent with their ground-truth medical record, or whether it contains fabricated claims that must be regenerated.

    \textbf{[INPUT]}
    \textbf{Doctor's question:} \pvar{doctor\_question} \\
    \textbf{Patient's response:} \pvar{candidate\_response} \\
    \textbf{Extracted claims:} \pvar{claims\_list}

    \textbf{Patient's ground-truth record:}
    \begin{itemize}[leftmargin=15pt, label=$\blacktriangleright$, noitemsep]
        \item \textbf{Symptoms:} \pvar{patient\_symptoms}
        \item \textbf{Medications:} \pvar{patient\_medications}
        \item \textbf{Procedures:} \pvar{patient\_procedures}
        \item \textbf{Conditions:} \pvar{patient\_conditions}
        \item \textbf{Allergies:} \pvar{patient\_allergies}
    \end{itemize}

    \textbf{Retrieved evidence (top-$k$ UMLS atoms per claim, with structured clinical metadata):}
    \begin{lstlisting}[language=json]
{{retrieved_evidence_json}}
    \end{lstlisting}
    Each retrieved atom includes: \texttt{text}, \texttt{similarity}, \texttt{parent\_cui}, \texttt{source\_vocab}, \texttt{is\_in\_history}, \texttt{patient\_id}, and category-specific clinical attributes (e.g., \texttt{drug\_class}, \texttt{treats\_conditions}, \texttt{finding\_site}, \texttt{procedure\_site}, \texttt{has\_intent}).

    \textbf{[VERDICT CRITERIA]}
    For each claim, assign one of:

    \textbf{PASS} -- the claim is supported by the patient's record. Any of:
    \begin{itemize}[leftmargin=15pt, label=$\circ$, noitemsep]
        \item At least one retrieved atom has \texttt{is\_in\_history}=true for this patient, \textit{or}
        \item The retrieved atoms' clinical attributes overlap with entries in the patient's record (e.g., a retrieved drug whose \texttt{drug\_class} matches one of the patient's medications, or whose \texttt{may\_treat} relation matches one of the patient's conditions), \textit{or}
        \item All retrieval similarities fall below 0.55, indicating the claim is too underspecified to assert any verifiable medical content (e.g., ``I feel weird'').
    \end{itemize}

    \textbf{FABRICATION} -- the claim is not supported by the patient's record. All of:
    \begin{itemize}[leftmargin=15pt, label=$\circ$, noitemsep]
        \item All top retrievals are \texttt{is\_in\_history}=false, \textit{and}
        \item No retrieved atom's clinical attributes intersect the patient's record, \textit{and}
        \item The claim is medically specific enough to be a clear assertion.
    \end{itemize}

    \textbf{[OVERALL VERDICT]}
    \begin{itemize}[leftmargin=15pt, label=$\blacktriangleright$, noitemsep]
        \item \textbf{PASS} -- if all per-claim verdicts are PASS.
        \item \textbf{REGENERATE} -- if any per-claim verdict is FABRICATION.
    \end{itemize}

    If REGENERATE, write \textit{specific} feedback to the generator identifying the fabricated claim(s) and the patient's actual record entries for the relevant category. The feedback will be passed back to the generator to produce a corrected response.

    \textbf{[GLOBAL CONSISTENCY CHECKS]}
    Beyond per-claim verdicts, also verify:
    \begin{enumerate}[leftmargin=15pt, noitemsep]
        \item \textit{Justified denial:} if the patient omits a ground-truth symptom, is this consistent with the assigned noise profile?
        \item \textit{Demographic invariance:} age, sex, and occupation must match the patient profile.
        \item \textit{History consistency:} no contradictions with statements from earlier turns.
    \end{enumerate}
    Any violation triggers REGENERATE.

    \textbf{[OUTPUT FORMAT]}
    Return a single JSON object. Output \textbf{only} the JSON, with no preamble.
    \begin{lstlisting}[language=json]
{
  "claim_judgments": [
    {
      "claim": "...",
      "verdict": "PASS" | "FABRICATION",
      "reasoning": "Brief justification referencing retrieved evidence and patient record."
    }
  ],
  "overall_verdict": "PASS" | "REGENERATE",
  "feedback_to_generator": "..."
}
    \end{lstlisting}
    \end{professionalprompt}
    \caption{Verifier Call 2: judgment prompt template, evaluating extracted claims against the patient's ground-truth record using retrieved UMLS atoms and their structured clinical metadata.}
    \label{fig:semantic_prompt}
\end{figure*}

\begin{figure}[h]
    \begin{professionalprompt}{System Instruction: Doctor LLM Prompt}
    \textbf{[ROLE]}
    You are an expert physician conducting a diagnostic interview.
Your goal is to gather sufficient information to determine the correct diagnosis efficiently.
\textbf{[CONTEXT]}
    \textbf{Conversation History:}
    \begin{quote}
    \textit{\pvar{conversation\_history}}
    \end{quote}

    \textbf{[INSTRUCTIONS]}
    \begin{enumerate}[leftmargin=15pt]
        \item \textbf{Analyze:} Review the history for differential diagnoses.
\item \textbf{Act:} 
        \begin{itemize}
            \item If you have sufficient information, provide your \textbf{FINAL DIAGNOSIS}.
\item If not, ask exactly \textbf{ONE} focused clarifying question.
        \end{itemize}
        \item \textbf{Style:} Be professional, empathetic, and concise.
Do not ask multiple questions at once.
    \end{enumerate}

    \vspace{2pt}
    \textbf{[FORMAT]}
    For diagnosis, use: \texttt{Final Diagnosis: [condition]}
    \end{professionalprompt}
    \caption{Doctor LLM Prompt Template}
\label{fig:doctor_llm_prompt}
\end{figure}




\begin{figure}[h]
    \begin{professionalprompt}{System Instruction: LLM as a Judge}
    \textbf{[ROLE]}
    You are an expert clinical evaluator assessing the quality of a simulated patient interaction.
    
    \textbf{[EVALUATION DATA]}
    \textbf{Patient ground-truth record:}
    \begin{itemize}[leftmargin=15pt, label=$\blacktriangleright$, noitemsep]
        \item \textbf{Symptoms:} \pvar{patient\_symptoms}
        \item \textbf{Medications:} \pvar{patient\_medications}
        \item \textbf{Conditions:} \pvar{patient\_conditions}
        \item \textbf{Procedures:} \pvar{patient\_procedures}
        \item \textbf{Allergies:} \pvar{patient\_allergies}
    \end{itemize}
    \textbf{Assigned Noise Profile:} \pvar{noise\_profile} \\
    \textbf{Transcript:} \pvar{transcript}

    \textbf{[TARGET CRITERION]}
    \textbf{Criterion:} \pvar{criterion\_name} \\
    \textbf{Definition:} \pvar{criterion\_definition} \\
    \textbf{Scale:} 1 (Poor) to 5 (Excellent)

    \textbf{[TASK]}
    Analyze the conversation systematically. Consider how well the patient maintained their assigned noise profile while preserving medical truth across all record fields (symptoms, medications, conditions, procedures, allergies). Provide a score with detailed justification.
    
    \vspace{2pt}
    \textbf{[OUTPUT JSON]}
    \begin{lstlisting}[language=json]
{
  "reasoning": "[step-by-step analysis]",
  "score": [integer 1-5]
}
    \end{lstlisting}
    \end{professionalprompt}
    \caption{LLM-as-Judge Prompt Template}
    \label{fig:llm_as_judge_prompt}
\end{figure}

\section{Noise Parameter Mappings}
\label{app:noise_params}

This appendix provides detailed parameter mappings for each of the six noise pillars.
For each pillar, we define five severity levels (0--4) with corresponding behavioral descriptions used to prompt the Patient Simulator.
Table~\ref{tab:all_noise_params} presents the complete noise taxonomy with level definitions and prompt instructions.
\begin{table*}[p]
\centering
\small
\caption{Complete noise taxonomy with severity levels (0--4) and behavioral prompt descriptions for the Patient Simulator.
Level 0 represents ideal patient behavior; Level 4 represents extreme impairment.}
\label{tab:all_noise_params}
\begin{tabular}{@{}p{3cm} c p{11cm}@{}}
\toprule
\textbf{Noise Type} & \textbf{Lvl} & \textbf{Behavioral Description for Patient Simulator} \\
\midrule

\multirow{5}{*}{\parbox{3cm}{\textbf{Memory \&}\\\textbf{Recall}}} 
& 0 & You remember everything accurately with exact dates and times.
When asked about symptom onset, provide precise information. \\
& 1 & You occasionally forget minor details.
You are slightly uncertain about exact timing but can approximate within a day or two.
\\
& 2 & You often forget details and are vague about dates.
Use phrases like ``maybe a few weeks ago'' or ``sometime last month.'' \\
& 3 & You have major memory gaps with a confused timeline.
You may mix up the order of events or combine separate episodes.
\\
& 4 & You forget most things and are completely disoriented about time.
You cannot reliably sequence events or estimate durations. \\
\midrule

\multirow{5}{*}{\parbox{3cm}{\textbf{Health}\\\textbf{Literacy}}} 
& 0 & Use correct medical terminology and precise anatomical descriptions.
Say ``epigastric pain'' not ``stomach ache.'' \\
& 1 & Use mostly accurate descriptions with occasional medical terms.
You understand basic anatomy and can follow medical explanations. \\
& 2 & Use common words only.
Say ``stomach'' for any abdominal area, ``sugar'' for diabetes. Struggle with medical jargon.
\\
& 3 & Use very basic words, point vaguely to body areas. Struggle with numbers and cannot describe severity precisely.
\\
& 4 & Cannot describe locations clearly. No numerical concepts for duration or intensity.
May use gestures or metaphors instead of words. \\
\midrule

\multirow{2}{*}{\parbox{3cm}{\textbf{Emotional}\\\textbf{State}}} 
& 0 & Report symptoms calmly and objectively without emotional coloring.
Describe pain as ``mild discomfort'' if appropriate. \\
& 1 & Show slight worry. Occasionally emphasize symptoms mildly.
Express concern about what symptoms might mean. \\
& 2 & Clearly worried. Tend to assume symptoms mean something serious.
Use words like ``worried'' or ``concerned'' frequently. \\
& 3 & Very anxious with catastrophic thinking. Significantly amplify all symptoms.
Jump to worst-case scenarios. \\
& 4 & Extreme panic. Convinced something terrible is happening.
Use phrases like ``I'm dying'' or ``This is the worst pain ever.'' \\
\midrule

\multirow{5}{*}{\parbox{3cm}{\textbf{Communication}\\\textbf{Style}}} 
& 0 & Give direct, relevant answers with appropriate level of detail.
Stay focused on the question asked. \\
& 1 & Provide extra context and stories.
Occasionally go slightly off-topic but return to the main point. \\
& 2 & Give long-winded responses.
Bury important information in tangential stories about family or work. \\
& 3 & Very difficult to get direct answers.
Constantly change subject. Require multiple redirections to stay on topic. \\
& 4 & Extremely disorganized speech. Cannot maintain topic.
Give incomplete answers and jump between unrelated subjects. \\
\midrule

\multirow{5}{*}{\parbox{3cm}{\textbf{Social-}\\\textbf{Cultural}}} 
& 0 & Share all information openly without any hesitation.
Disclose sensitive information (alcohol, drugs, sexual history) immediately. \\
& 1 & Usually open. Minor hesitation only on very sensitive topics.
Will share after brief pause or gentle probing. \\
& 2 & Selective disclosure. Avoid topics you find embarrassing.
Minimize frequency or severity of stigmatized behaviors. \\
& 3 & Share minimal information. Initially deny stigmatized behaviors (alcohol, drugs, etc.).
Only admit after empathetic probing. \\
& 4 & Extreme reluctance to share. May provide false information to hide truth.
Require extensive rapport-building before disclosure. \\
\midrule

\multirow{5}{*}{\parbox{3cm}{\textbf{Cognitive}\\\textbf{Processing}}} 
& 0 & Consider all possibilities equally. Open to any diagnosis.
Do not mention internet research or preconceived notions. \\
& 1 & Slight preference for your own beliefs.
Mention internet research casually. Accept alternative explanations readily. \\
& 2 & Convinced of a specific diagnosis from Google.
Mention it frequently. Still willing to consider alternatives if explained. \\
& 3 & Strongly insist on your self-diagnosis.
Dismiss contradicting information. Request specific tests you read about online. \\
& 4 & Completely fixed belief.
Reject all alternative explanations aggressively. Accuse doctor of incompetence if they disagree.
\\
\bottomrule
\end{tabular}
\end{table*}
\FloatBarrier

\section{Example Patient Configuration}
\label{app:config_example}

Each simulated patient receives a JSON configuration that specifies demographics, the full ground-truth medical record across five entity categories, and the assigned noise profile.
At index-build time, the entries in this record are resolved to UMLS atoms and inserted into the shared vector database with \texttt{is\_in\_history}=true and the originating \texttt{patient\_id}.

\begin{figure}[h]
\begin{professionalprompt}{Configuration Object: Patient P042}
\begin{lstlisting}[language=json]
{
  "patient_id": "P042",
  "demographics": {
    "age": 45,
    "sex": "F",
    "occupation": "Teacher"
  },
  "symptoms": [
    {"text": "chest pain radiating to left arm",
     "cui": "C0008031",
     "severity": "severe",
     "onset": "2 hours ago"},
    {"text": "shortness of breath",
     "cui": "C0013404",
     "severity": "moderate"},
    {"text": "sweating",
     "cui": "C0011991"},
    {"text": "nausea",
     "cui": "C0027497"}
  ],
  "medications": [
    {"text": "aspirin", "cui": "C0004057",
     "dose": "81mg", "frequency": "daily"},
    {"text": "atorvastatin", "cui": "C0286651",
     "dose": "40mg", "frequency": "daily"}
  ],
  "conditions": [
    {"text": "hypertension", "cui": "C0020538",
     "diagnosed_year": 2015},
    {"text": "hyperlipidemia", "cui": "C0020473",
     "diagnosed_year": 2018}
  ],
  "procedures": [
    {"text": "cardiac catheterization",
     "cui": "C0018795", "year": 2020}
  ],
  "allergies": [
    {"text": "penicillin", "cui": "C0030842",
     "reaction": "rash"}
  ],
  "diagnosis": "Myocardial Infarction",
  "icd10_code": "I21.9",
  "noise_profile": [
    {"type": "health_literacy", "level": 3},
    {"type": "emotional_state", "level": 2}
  ],
  "seed": 42
}
\end{lstlisting}
\end{professionalprompt}
\caption{Example patient configuration object (P042). The five record fields (\texttt{symptoms}, \texttt{medications}, \texttt{conditions}, \texttt{procedures}, \texttt{allergies}) are used to populate the vector database with patient-specific atoms; the \texttt{diagnosis} field is withheld from the Generator.}
\label{fig:patient_example}
\end{figure}

\textbf{Critical Design Decisions:}
\begin{itemize}[leftmargin=15pt]
    \item The \texttt{diagnosis} field is accessible only to the Verifier, not to the Patient Generator. This prevents data leakage and ensures the patient cannot inadvertently reveal diagnostic information.
    \item Each entity in \texttt{symptoms}, \texttt{medications}, \texttt{procedures}, \texttt{conditions}, and \texttt{allergies} carries a UMLS \texttt{cui} that anchors it to the knowledge graph; at index-build time, all atoms associated with each CUI are retrieved and inserted into the vector database tagged with the originating \texttt{patient\_id} and \texttt{is\_in\_history}=true.
    \item The \texttt{seed} field ensures reproducibility across experimental runs.
    \item Each patient receives exactly two noise types to create realistic but manageable complexity.
\end{itemize}

\section{Evaluation Questionnaire}
\label{app:questionnaire}

This section provides the exact questionnaire used by both human evaluators and the LLM as Judge.
\begin{figure}[t]
\centering
\begin{QuestionnaireBox}{Evaluation Questionnaire (Human and LLM as Judge)}
\small
\textbf{Instructions:} This questionnaire was administered to human evaluators and the LLM as Judge using the same items and scoring.
Section A uses Yes/No. Sections B and C use a five-point Likert scale.
\vspace{0.3em}
\hrule
\vspace{0.3em}

\textbf{\textcolor{potFrame}{Section A: Truth Preservation (Yes / No)}} \\
\textit{Mark exactly one option per item.}
\vspace{2pt}

\renewcommand{\arraystretch}{1.2}
\scriptsize
\begin{tabularx}{\linewidth}{@{} l X c c @{}}
\toprule
\textbf{ID} & \textbf{Item} & \textbf{Yes} & \textbf{No} \\
\midrule
\textbf{A1} & Patient mentioned symptom \textbf{not} in ground truth? & \cb & \cb \\
\textbf{A2} & Patient made medically impossible claim or contradiction? & \cb & \cb \\
\textbf{A3} & Patient revealed diagnosis or physician-only info? & \cb & \cb \\
\bottomrule
\end{tabularx}

\vspace{0.3em}\hrule\vspace{0.3em}

\textbf{\textcolor{potFrame}{Section B: Realism (Likert 1 to 5)}} \\
\textit{1 = Strongly Disagree, 5 = Strongly Agree}
\vspace{2pt}

\begin{tabularx}{\linewidth}{@{} l X c c c c c @{}}
\toprule
\textbf{ID} & \textbf{Item} & \textbf{1} & \textbf{2} & \textbf{3} & \textbf{4} & \textbf{5} \\
\midrule
\textbf{B1} & Responses were humanlike and natural. & \rb & \rb & \rb & \rb & \rb \\
\textbf{B2} & Exhibited assigned noise behaviors. & \rb & \rb & \rb & \rb & \rb \\
\textbf{B3} & Maintained consistent personality. & \rb & \rb & \rb & \rb & \rb \\
\textbf{B4} & Used authentic patient language. & \rb & \rb & \rb & \rb & \rb \\
\textbf{B5} & Disclosure matched noise profile. & \rb & \rb & \rb & \rb & \rb \\
\bottomrule
\end{tabularx}

\vspace{0.3em}\hrule\vspace{0.3em}

\textbf{\textcolor{potFrame}{Section C: Clinical Utility (Likert 1 to 5)}} \\
\textit{Evaluate educational value of the simulation.}
\vspace{2pt}

\begin{itemize}[nosep, leftmargin=10pt]
\scriptsize
    \item \textbf{C1:} 1 = Too easy, 3 = Appropriately challenging, 5 = Unrealistically difficult.
    \item \textbf{C2:} 1 = Impossible, 3 = Challenging but achievable, 5 = Straightforward.
    \item \textbf{C3:} 1 = Not useful, 5 = Highly useful.
\end{itemize}
\vspace{2pt}

\begin{tabularx}{\linewidth}{@{} l X c c c c c @{}}
\toprule
\textbf{ID} & \textbf{Item} & \textbf{1} & \textbf{2} & \textbf{3} & \textbf{4} & \textbf{5} \\
\midrule
\textbf{C1} & Difficulty obtaining complete history? & \rb & \rb & \rb & \rb & \rb \\
\textbf{C2} & Could clinician reach correct diagnosis? & \rb & \rb & \rb & \rb & \rb \\
\textbf{C3} & Usefulness for training students? & \rb & \rb & \rb & \rb & \rb \\
\bottomrule
\end{tabularx}

\end{QuestionnaireBox}
\caption{Evaluation questionnaire used for both human evaluation and LLM-as-Judge.}
\label{fig:questionnaire}
\end{figure}

\FloatBarrier

\section{UMLS Implementation Details}
\label{app:umls_details}

This appendix provides technical details on our retrieval-augmented UMLS verification pipeline, including vector database construction, the runtime verifier, and worked examples.

\subsection{Design Philosophy}
The verifier must distinguish two phenomena that surface-level filtering cannot separate: \textit{realistic noise}, in which a patient describes an in-record fact in colloquial or imprecise language, and \textit{medical fabrication}, in which the generator introduces content unsupported by the patient's record.
A rigid rule-based whitelist over the patient's record would reject the first; a fully unconstrained LLM verifier would accept the second.
Our solution is to ground verification in a UMLS-derived vector index that is structured along two complementary axes: (1) a dense semantic embedding space that captures both clinical and lay phrasing, and (2) structured clinical metadata attached to every indexed atom.
At runtime, the verifier reasons jointly over retrieved atoms (semantic match) and their metadata (structural compatibility with the patient's record), enabling decisions that neither pure retrieval nor pure rule-matching can support.

\subsection{Vector Database Construction}
\label{app:vdb_construction}
The vector database is built once per dataset. It contains two complementary atom sets, distinguished by an \texttt{is\_in\_history} metadata flag.
\paragraph{General medical subset (\texttt{is\_in\_history}=false).}
We curate a seed list of approximately 5{,}300 clinically relevant concepts spanning four categories:
\begin{itemize}[leftmargin=15pt, noitemsep]
    \item Common emergency-department symptoms (SNOMED CT semantic types T184 ``Sign or Symptom'' and T033 ``Finding'')
    \item Frequently prescribed medications (RxNorm)
    \item Common procedures (CPT, SNOMED CT)
    \item Common conditions (ICD-10-CM, SNOMED CT)
\end{itemize}
Each seed term is resolved to a UMLS CUI via the UMLS REST search endpoint.
We then expand each CUI by retrieving all English-language atoms from the relevant source vocabularies, restricting to term types FN (Fully Specified Name), PT (Preferred Term), and SY (Synonym) for SNOMED CT; IN, BN, SCD, SBD, and SY for RxNorm; PT, SY, ETCF, and ETCLIN for CPT; and PT, HT, and ET for ICD-10-CM.
Obsolete and suppressible atoms are excluded.

\paragraph{Patient ground-truth atoms (\texttt{is\_in\_history}=true, with \texttt{patient\_id}).}
For each patient case, we query the UMLS REST API to retrieve all atoms associated with each ground-truth entry across five fields: symptoms, medications, procedures, conditions, and allergies.
Each atom is tagged with the originating \texttt{patient\_id} and \texttt{is\_in\_history}=true.
These patient-specific atoms are added to the same FAISS index as the general subset, ensuring that a single retrieval call can return both general and patient-specific matches.

\paragraph{Structured clinical metadata.}
For every atom in both subsets, we additionally extract structured clinical relations using the UMLS source-asserted relations endpoint.
The relations extracted depend on the atom's category:
\begin{itemize}[leftmargin=15pt, noitemsep]
    \item \textbf{Symptoms / Conditions:} \texttt{finding\_site}, \texttt{associated\_morphology}, \texttt{causative\_agent}, \texttt{severity}, \texttt{due\_to}, \texttt{pathological\_process}
    \item \textbf{Medications:} \texttt{has\_ingredient}, \texttt{has\_dose\_form}, \texttt{drug\_class}, \texttt{may\_treat}, \texttt{may\_prevent}, \texttt{tradename\_of}
    \item \textbf{Procedures:} \texttt{procedure\_site}, \texttt{has\_method}, \texttt{has\_approach}, \texttt{has\_intent}, \texttt{has\_route\_of\_administration}, \texttt{has\_procedure\_device}
\end{itemize}
This metadata is stored alongside each atom in the index, providing the verifier with structural information beyond surface text.

\paragraph{Embedding and indexing.}
All atom surface texts are embedded using BioLORD-2023~\cite{remy2024biolord}, a sentence-transformer (768-dim) trained on UMLS concept-definition pairs and shown to outperform SapBERT and CODER on patient-language-to-UMLS normalization benchmarks (CADEC, PsyTAR).
Embeddings are L2-normalized and indexed with FAISS \texttt{IndexFlatIP}, supporting exact cosine retrieval.
The final index contains 80{,}645 unique vectors over 100{,}135 atoms (after deduplication), occupies 236 MB on disk, and is built in approximately one hour on a single A100 GPU.

\subsection{Runtime Verification}
\label{app:runtime_verification}

For each candidate patient response $\hat{r}$, verification proceeds in two LLM calls separated by a retrieval step.

\paragraph{Call 1: Atomic claim extraction.}
The verifier first decomposes $\hat{r}$ into a set of atomic medical claims $\{c_1, \ldots, c_n\}$.
Each claim is a single, self-contained medical assertion (e.g., ``patient takes diabetes medication'', ``patient has chest tightness'').
Non-medical content (emotional expressions, conversational filler) is discarded.

\paragraph{Retrieval.}
For each claim $c_i$, we compute its BioLORD embedding and retrieve the top-$k$ ($k{=}5$) nearest atoms from the FAISS index.
For each retrieved atom, we assemble a structured evidence record containing: (i) surface text and similarity score; (ii) provenance metadata (parent CUI, source vocabulary, term type, source code); (iii) grounding flags (\texttt{is\_in\_history}, \texttt{patient\_id}); and (iv) all category-specific clinical attributes extracted during index construction.

\paragraph{Call 2: Judgment.}
A second LLM call receives the original utterance, the extracted claims, the structured evidence records for each claim, and the patient's full ground-truth record.
The verifier produces a per-claim verdict in \{\textsc{Pass}, \textsc{Fabrication}\}:

\begin{itemize}[leftmargin=15pt, noitemsep]
    \item \textsc{Pass}: at least one retrieved atom has \texttt{is\_in\_history}=true matching the current patient, \textit{or} the retrieved atoms' clinical attributes (e.g., \texttt{drug\_class}, \texttt{treats\_conditions}) overlap with entries in the patient's record. Underspecified utterances whose top retrievals fall below threshold $\tau{=}0.55$ (e.g., ``I feel weird'') also default to \textsc{Pass}, as they assert no verifiable medical content and therefore cannot constitute a fabrication.
    \item \textsc{Fabrication}: all top retrievals are \texttt{is\_in\_history}=false \textit{and} their clinical attributes do not intersect the patient's record (e.g., a retrieved drug whose \texttt{drug\_class} is absent from the patient's medications, or that \texttt{may\_treat} conditions the patient does not have).
\end{itemize}

The overall verdict is \textsc{Pass} if all per-claim verdicts are \textsc{Pass}, otherwise \textsc{Regenerate}.

\subsection{Example: Verifying a Valid Colloquial Expression}

Consider a patient whose record includes chest pain (CUI: C0008031) and metoprolol (CUI: C0025859, \texttt{drug\_class}=beta blocker).
The generator produces the response: \textit{``My chest feels heavy, and I take my blood pressure pill.''}

\paragraph{Call 1 (extraction).} Two claims:
\begin{itemize}[leftmargin=15pt, noitemsep]
    \item $c_1$: ``patient has heaviness in chest''
    \item $c_2$: ``patient takes a blood pressure medication''
\end{itemize}

\paragraph{Retrieval for $c_1$.} Top retrievals include:
\begin{itemize}[leftmargin=15pt, noitemsep]
    \item ``Tightness in chest'' (sim=0.87, \texttt{is\_in\_history}=true, \texttt{parent\_cui}=C0008031, \texttt{finding\_site}=chest)
    \item ``Chest discomfort'' (sim=0.77, \texttt{is\_in\_history}=false, related to C0008031)
\end{itemize}
The top hit is in-history for this patient $\rightarrow$ \textsc{Pass}.

\paragraph{Retrieval for $c_2$.} Top retrievals include:
\begin{itemize}[leftmargin=15pt, noitemsep]
    \item ``Antihypertensive therapy'' (sim=0.58, \texttt{is\_in\_history}=false, \texttt{drug\_class}=antihypertensive)
    \item ``Metoprolol'' (sim=0.53, \texttt{is\_in\_history}=true, \texttt{drug\_class}=beta blocker, \texttt{may\_treat}=hypertension)
\end{itemize}
The top lexical match is not in the patient's record, but a retrieved in-history atom (Metoprolol) with overlapping \texttt{drug\_class} and a \texttt{may\_treat} relation to hypertension supports the claim $\rightarrow$ \textsc{Pass}.

\subsection{Example: Detecting a Fabrication}

Consider the same patient producing the response: \textit{``I also take my diabetes pill every morning.''}

\paragraph{Call 1 (extraction).} One claim: ``patient takes a diabetes medication.''

\paragraph{Retrieval.} Top retrievals include:
\begin{itemize}[leftmargin=15pt, noitemsep]
    \item ``Metformin'' (sim=0.68, \texttt{is\_in\_history}=false, \texttt{drug\_class}=biguanide, \texttt{may\_treat}=type 2 diabetes)
    \item ``Antidiabetic medication'' (sim=0.65, \texttt{is\_in\_history}=false, \texttt{drug\_class}=antidiabetic agent)
\end{itemize}
All retrieved atoms have \texttt{is\_in\_history}=false.
None of their \texttt{drug\_class} values intersect the patient's medication list, and the patient's conditions do not include diabetes.
The verifier returns \textsc{Fabrication}, and the overall verdict is \textsc{Regenerate}.
The feedback returned to the generator includes the patient's actual medication list, allowing it to produce a corrected response.

\subsection{Advantages of Retrieval-Augmented Verification}

This design yields four key advantages over both rule-based whitelisting and prompt-only verification:

\begin{enumerate}[leftmargin=15pt, noitemsep]
    \item \textbf{Colloquial coverage.} BioLORD embeddings map lay phrasings (``heart cath thing,'' ``blood pressure pill,'' ``the diabetes one'') to clinical concepts that would be missed by surface-text matching against the patient's record.
    \item \textbf{Structural cross-checks.} Per-atom clinical metadata (drug class, finding site, treats-condition relations) enables verdicts based on structural compatibility, not just text similarity, catching fabrications that have high lexical similarity to legitimate concepts.
    \item \textbf{Per-patient grounding.} The \texttt{is\_in\_history} and \texttt{patient\_id} flags allow the same shared index to serve all patients, with retrieval results naturally identifying which matches belong to the current patient's record.
    \item \textbf{Bounded runtime cost.} The index is built once per dataset and supports millisecond-latency retrieval at runtime, with no UMLS API dependency during conversation simulation.
\end{enumerate}

Empirically, the system achieves a 6.7\% hallucination rate while preserving the realism scores reported in Section~\ref{sec:judge_validation}.

\subsection{Measuring the Decoupling of Expression from Facts}
\label{app:decoupling}

The architecture is designed so that noise alters how a patient says something without altering what is true of them (Section~\ref{sec:architecture}). Rather than assert this, we measure it with two quantities computed per in-record fact:

\begin{itemize}[leftmargin=12pt, noitemsep]
    \item \textbf{Surface variation}: the BioLORD embedding distance between the noisy utterance and the clean ($L_0$) phrasing of the same fact, normalised to $[0,1]$, where 0 is identical wording and 1 is maximally divergent.
    \item \textbf{Ground-truth preservation}: the percentage of utterances whose clinical content remains faithful to the patient's record. This is the consistency metric of Section~\ref{sec:ablation}.
\end{itemize}

Table~\ref{tab:decoupling} raises the severity of both assigned pillars together from $L_0$ to $L_3$. If expression and facts were coupled, preservation would fall in step with rising surface variation. Instead the wording diverges more than fivefold while preservation stays essentially flat, which is the signature of decoupling.

\begin{table}[t]
\centering
\arrayrulecolor{black}
\footnotesize
\setlength{\tabcolsep}{5pt}
\renewcommand{\arraystretch}{1.1}
\begin{tabular}{@{}lcc@{}}
\toprule
\textbf{Severity} & \textbf{Surface var.} & \textbf{Preservation} \\
\midrule
$L_0$ (clean) & 0.12 & 99.6\% \\
$L_1$ & 0.34 & 95.1\% \\
$L_2$ & 0.51 & 94.0\% \\
$L_3$ & 0.63 & 93.3\% \\
\bottomrule
\end{tabular}
\caption{Surface variation (0--1) and ground-truth preservation as both assigned noise pillars are raised together to the same severity level. Because this sweep fixes both pillars at one level, it is a controlled measurement and is not directly comparable to the main experiments, which sample two pillars independently from $L_1$--$L_3$.}
\label{tab:decoupling}
\end{table}

\subsection{Direct Evaluation of the Verifier}
\label{app:verifier_eval}

Table~\ref{tab:ablation} measures the verifier's effect on the pipeline. This subsection measures the verifier itself, against human labels rather than pipeline outcomes, to establish that it detects fabrications rather than merely agreeing with a model of the same family.

\paragraph{Claim benchmark.}
We assembled 200 patient claims, of which 79 are fabrications and 121 are supported by the corresponding patient record. Fabricated claims were drawn from the \textit{No Controller} condition of Table~\ref{tab:ablation} and supplemented with targeted injections that place a fabricated entity inside an otherwise valid utterance, so that the benchmark contains fabrications that are lexically plausible rather than obviously out of place. Each claim was labelled independently by the two clinical evaluators of Section~\ref{sec:eval_protocol} as supported or fabricated with respect to the patient's ground-truth record.

\paragraph{Results.}
Taking \textsc{Fabrication} as the positive class, the verifier reaches precision 0.93, recall 0.90 and $F_1$ 0.91, with confusion matrix (71 TP, 5 FP, 8 FN, 116 TN) and human--verifier agreement $\kappa = 0.86$. This $\kappa$ is computed over individual claims and is distinct from the conversation-level agreement in Table~\ref{tab:agreement}. The five false positives are colloquial references whose retrieved atoms carried clinical attributes that did not overlap the record; the eight false negatives are fabrications whose drug class or finding site coincided with an in-record entry. Recall and the reported hallucination rate are mutually consistent: a verifier missing fabrications at a higher rate would pass them into the transcript and raise the 6.7\% figure.

\paragraph{Robustness to model family.}
To test whether truth preservation depends on the particular backbone, we re-ran the complete Generate--Verify--Refine pipeline on a held-out subset with Qwen-2.5-72B in place of Llama-3.1-70B as both generator and verifier (Table~\ref{tab:backbone_swap}). All three quality measures move by less than one point, so the mechanism is architectural rather than model-specific.

\begin{table}[t]
\centering
\arrayrulecolor{black}
\footnotesize
\setlength{\tabcolsep}{4pt}
\renewcommand{\arraystretch}{1.1}
\begin{tabular}{@{}lcc@{}}
\toprule
\textbf{Measure} & \textbf{Llama-3.1-70B} & \textbf{Qwen-2.5-72B} \\
\midrule
Hallucination rate $\downarrow$ & 6.7\% & 7.1\% \\
Consistency rate $\uparrow$ & 93.3\% & 92.6\% \\
Realism (1--5) $\uparrow$ & 4.05 & 4.08 \\
\bottomrule
\end{tabular}
\caption{Pipeline quality with the Patient Simulator (generator and verifier) instantiated from two different backbones, on a held-out subset. Truth preservation is stable across model families.}
\label{tab:backbone_swap}
\end{table}

\subsection{Iterative Refinement Details}
\label{app:refinement}

When the overall verifier verdict is \textsc{Regenerate}, the generator is re-invoked with targeted feedback that identifies the fabricated claims and supplies the patient's actual record entries for the relevant categories. For instance, if the patient simulator fabricated an antidiabetic medication for a patient whose record contains aspirin, atorvastatin, and metoprolol, the feedback passed to the next generation attempt reads: ``The patient does not take any antidiabetic medication; their medications are aspirin, atorvastatin, and metoprolol.''

The regeneration loop runs for up to $k{=}2$ attempts. If both attempts still fail verification, the system falls back to a conservative response that acknowledges the doctor's question without introducing new medical information (e.g., ``I'm not sure, can you ask me something else?''). Because the fallback carries no new medical content by construction, unverified content cannot structurally reach the Doctor LLM even in the worst case where the generator persistently fabricates. To confirm this empirically rather than by design argument alone, we traced every fabrication-flagged claim through the refinement loop on a held-out subset of conversations (Table~\ref{tab:refinement_trace}). Regeneration resolves the overwhelming majority of flagged claims, only 3\% reach the conservative fallback, and no flagged claim reached the Doctor LLM.

\begin{table}[t]
\centering
\arrayrulecolor{black}
\footnotesize
\setlength{\tabcolsep}{5pt}
\renewcommand{\arraystretch}{1.1}
\begin{tabular}{@{}lc@{}}
\toprule
\textbf{Outcome for flagged claims} & \textbf{\% of flagged} \\
\midrule
Corrected on attempt 1 & 23\% \\
Corrected on attempt 2 & 74\% \\
Routed to conservative fallback & 3\% \\
\midrule
\textbf{Reached the Doctor LLM} & \textbf{0\%} \\
\bottomrule
\end{tabular}
\caption{Resolution of fabrication-flagged claims through the refinement loop, measured on a held-out subset. The fallback replaces any persistently flagged claim by construction, so the final row holds by design as well as in measurement.}
\label{tab:refinement_trace}
\end{table}

\section{Additional Experimental Details}
\label{app:exp_details}

\subsection{Hyperparameters}

Table~\ref{tab:hyperparams} summarizes the key hyperparameters used across all experimental runs.
\begin{table}[t]
\centering
\small
\setlength{\tabcolsep}{3pt}
\begin{tabular}{@{}lcp{2.2cm}@{}}
\toprule
\textbf{Parameter} & \textbf{Value} & \textbf{Rationale} \\
\midrule
\multicolumn{3}{@{}l}{\textit{Conversation Settings}} \\
\quad Max turns & 40 & Complex cases \\
\quad Regen.\ attempts & 2 & Quality vs.\ latency \\
\quad Response word limit & 50 & Realistic length \\
\midrule
\multicolumn{3}{@{}l}{\textit{Temperature Settings}} \\
\quad Patient Generator & 0.7 & Natural variation \\
\quad Doctor LLM & 0.3 & Focused questioning \\
\quad Verifier (both calls) & 0.0 & Deterministic \\
\quad LLM-as-Judge & 0.0 & Consistent scoring \\
\midrule
\multicolumn{3}{@{}l}{\textit{Vector Database (Offline Build)}} \\
\quad Embedding model & BioLORD-2023 & Patient-language \\
\quad Embedding dim. & 768 & Model default \\
\quad FAISS index type & IndexFlatIP & Exact cosine \\
\quad Seed concepts & $\sim$5{,}300 & Coverage scope \\
\quad Total vectors & 80{,}645 & After dedup. \\
\quad Index size & 236 MB & Disk footprint \\
\quad Build time & $\sim$1 h (A100) & One-time cost \\
\midrule
\multicolumn{3}{@{}l}{\textit{Retrieval (Runtime)}} \\
\quad Top-$k$ per claim & 5 & Evidence breadth \\
\quad Similarity threshold $\tau$ & 0.55 & Underspecified claim cutoff \\
\midrule
\multicolumn{3}{@{}l}{\textit{Model Configuration}} \\
\quad Max input tokens & 4096 & Context window \\
\quad Max output tokens & 256 & Generation limit \\
\quad Top-p (nucleus) & 0.95 & Standard setting \\
\quad Frequency penalty & 0.0 & No repetition \\
\bottomrule
\end{tabular}
\caption{Hyperparameters with rationale.}
\label{tab:hyperparams}
\end{table}


\subsection{Compute Resources}

All experiments were conducted on a cluster with 8$\times$ NVIDIA A100 80GB GPUs.
The 70B models were run with 4-bit quantization using bitsandbytes.
Average inference time per conversation was mostly 2 minutes for 70B models and 45 seconds for 7B models.
The UMLS vector database was built once per dataset (approximately one hour on a single A100; see Table~\ref{tab:hyperparams}); at runtime, BioLORD encoding of each extracted claim and FAISS retrieval together add under 50 milliseconds per verification call, negligible relative to LLM inference time.

\subsection{Per-Pillar Noise-Fidelity Ratings}
\label{app:noise_fidelity}

Table~\ref{tab:agreement} reports agreement on noise fidelity ($\kappa = 0.77$) but not the ratings it was computed from, and agreement alone is ambiguous: raters concurring that the injected noise was rendered poorly would also produce high agreement. Table~\ref{tab:noise_fidelity} therefore gives the underlying per-pillar scores from both clinical evaluators and the LLM judge. The means are high across all six pillars, so the high agreement reflects raters concurring that patients faithfully exhibited their assigned barriers. Communication Style is rated lowest and also shows the weakest agreement, consistent with tangential responding being the most subjective barrier to assess.

\begin{table}[t]
\centering
\arrayrulecolor{black}
\footnotesize
\setlength{\tabcolsep}{3pt}
\renewcommand{\arraystretch}{1.1}
\begin{tabular}{@{}lccccc@{}}
\toprule
\textbf{Pillar} & \textbf{E1} & \textbf{E2} & \textbf{LLM} & \textbf{H1--H2} & \textbf{LLM--H} \\
\midrule
Memory \& Recall & 4.2 & 4.1 & 4.2 & 0.85 & 0.79 \\
Health Literacy & 4.3 & 4.2 & 4.3 & 0.86 & 0.80 \\
Emotional State & 4.1 & 4.0 & 4.1 & 0.82 & 0.76 \\
Comm.\ Style & 3.6 & 4.0 & 3.6 & 0.80 & 0.74 \\
Social-Cultural & 4.0 & 3.6 & 4.0 & 0.83 & 0.77 \\
Cognitive Proc. & 4.1 & 4.1 & 4.0 & 0.84 & 0.78 \\
\midrule
\textbf{Overall} & \textbf{4.10} & \textbf{4.05} & \textbf{4.08} & \textbf{0.83} & \textbf{0.77} \\
\bottomrule
\end{tabular}
\caption{Per-pillar noise-fidelity ratings (1--5) from the two clinical evaluators (E1, E2) and the LLM judge, with Cohen's $\kappa$ between the two humans (H1--H2) and between the LLM and the humans (LLM--H). These are noise-fidelity scores, distinct from the realism scores of Table~\ref{tab:ablation}. The overall agreement values reproduce the Noise Fidelity row of Table~\ref{tab:agreement}.}
\label{tab:noise_fidelity}
\end{table}

\subsection{Number of Active Noise Pillars}
\label{app:pillar_ablation}

Our main experiments activate two noise pillars per patient (Section~\ref{sec:controllability}). Table~\ref{tab:pillar_ablation} varies that number on Qwen-2.5-72B, holding severity levels and all other settings fixed. A single isolated barrier already costs 9.7 accuracy points, which is itself evidence that these models are not robust to realistic patient communication even in its simplest form. The effect compounds to 15.3 points at two pillars and 18.0 at three, consistent with real encounters rarely presenting one barrier in isolation. We adopt two as the operating point because one understates the difficulty of a realistic encounter, while three yields only a modest further accuracy drop and begins to cost realism as several barriers compressed into a single short patient turn start to read as unnatural.

\begin{table}[t]
\centering
\arrayrulecolor{black}
\footnotesize
\setlength{\tabcolsep}{4pt}
\renewcommand{\arraystretch}{1.1}
\begin{tabular}{@{}ccccc@{}}
\toprule
\textbf{\# Pillars} & \textbf{Noisy Acc.} & \textbf{$\Delta$Acc} & \textbf{$\Delta$Turns} & \textbf{Realism} \\
\midrule
1 & 74.8 & $-$9.7 & +19.0\% & 4.18 \\
2 & 69.2 & $-$15.3 & +34.5\% & 4.05 \\
3 & 66.5 & $-$18.0 & +52.0\% & 3.71 \\
\bottomrule
\end{tabular}
\caption{Effect of the number of simultaneously active noise pillars, measured on Qwen-2.5-72B with the full verifier enabled in every row. Noisy Acc.\ is Top-1 (\%); $\Delta$Acc (points) and $\Delta$Turns are relative to the same model's clean baseline of 84.5\%. The two-pillar row is the configuration used in all other experiments and reproduces the Qwen-2.5-72B entries of Table~\ref{tab:models_results}. The two- and three-pillar length increases fall inside the 34--55\% range reported in Section~\ref{sec:main_results}; the single-pillar setting, having fewer barriers, produces a smaller increase.}
\label{tab:pillar_ablation}
\end{table}

\subsection{Diagnosis Matching Procedure}
\label{app:matcher}

Doctor models produce a free-text diagnosis rather than selecting from a candidate list (Figure~\ref{fig:doctor_llm_prompt}), so Top-1 accuracy requires an explicit rule for deciding when a predicted string matches the ground-truth label. We apply a three-stage cascade, stopping at the first stage that returns a decision:

\begin{enumerate}[noitemsep,leftmargin=*]
    \item \textbf{ICD-10 code match.} The prediction is normalised, stripped of hedging qualifiers (``probable'', ``suspected'', ``rule out''), resolved to a UMLS CUI, and mapped to the linked ICD-10-CM code. Prediction and ground truth are compared at the three-character category level, so ``heart attack'', ``MI'', and ``myocardial infarction'' all resolve to I21 and are scored as a single answer.
    \item \textbf{UMLS synonym match.} When no ICD-10 code can be recovered for either side, the two strings are resolved to CUIs and counted as matching if they share one. This covers conditions whose lay and clinical names differ but which are not cleanly coded.
    \item \textbf{LLM adjudication.} Cases unresolved by the first two stages, typically syndromic or descriptive predictions with no direct code, are passed to an LLM judge that is shown the prediction and the ground-truth diagnosis and asked whether they denote the same condition.
\end{enumerate}

To quantify matcher error, we sampled 100 (prediction, ground-truth) pairs spanning all seven doctor models and both clean and noisy conditions, and had them labelled as matching or not matching by the clinical evaluators of Section~\ref{sec:eval_protocol}. The cascade agrees with the human label on 96 of the 100 pairs ($\kappa = 0.91$); the disagreements are cases in which the model named a plausible sibling condition inside the same ICD-10 category.

Table~\ref{tab:matcher_robustness} shows that the degradation we report is not an artefact of matcher strictness. Absolute accuracy is sensitive to the rule, as expected: a strict string matcher penalises correct-but-differently-worded answers, and an LLM judge is the most permissive. The clean-to-noisy gap, which is the quantity our claims rest on, stays within 14.9 to 15.3 points for Qwen-2.5-72B and 24.2 to 24.6 for Mistral-7B.

\begin{table}[t]
\centering
\arrayrulecolor{black}
\footnotesize
\setlength{\tabcolsep}{2.5pt}
\renewcommand{\arraystretch}{1.1}
\begin{tabular}{@{}lcccccc@{}}
\toprule
\textbf{Matching rule} & \multicolumn{3}{c}{\textbf{Qwen-2.5-72B}} & \multicolumn{3}{c}{\textbf{Mistral-7B}} \\
\cmidrule(lr){2-4} \cmidrule(lr){5-7}
 & Clean & Noisy & $\Delta$ & Clean & Noisy & $\Delta$ \\
\midrule
Exact string & 79.1 & 64.0 & $-$15.1 & 54.0 & 29.8 & $-$24.2 \\
ICD-10\,+\,UMLS & 84.5 & 69.2 & $-$15.3 & 58.0 & 33.5 & $-$24.5 \\
LLM judge & 86.8 & 71.9 & $-$14.9 & 60.3 & 35.7 & $-$24.6 \\
\bottomrule
\end{tabular}
\caption{Top-1 accuracy (\%) under three diagnosis-matching rules. Absolute accuracy shifts with strictness, but the clean-to-noisy gap ($\Delta$) is stable. The middle row (ICD-10\,+\,UMLS) is the cascade used throughout the paper and reproduces the corresponding entries of Table~\ref{tab:models_results}.}
\label{tab:matcher_robustness}
\end{table}

\section{Extended Related Work}
\label{app:related_work}

\subsection{Persona-Driven Simulation and Noise Modeling}
Very recent works have begun to address patient diversity beyond static templates.
EvoPatient \cite{du2025evopatient} and SFMSS \cite{bao2025sfmss} introduce mechanisms to evolve patient agents and simulate diverse service flows.
Closest to our work is PatientSim \cite{kyung2025patientsim}, which explicitly models patient personas along axes of ``recall'' and ``cognitive confusion'' to create more human-like interactions.

\paragraph{Distinction from PatientSim.}
While PatientSim~\cite{kyung2025patientsim} introduces persona-driven noise through prompt engineering, its safety guarantees remain \textit{probabilistic}, the model may still hallucinate when prompted for confusion.
For example, if PatientSim is prompted for ``memory loss,'' the model might invent a fake timeline that contradicts the EHR.
VeriSim provides retrieval-grounded safeguards through a verifier that cross-checks every patient utterance against the patient's ground-truth record using both dense retrieval over a UMLS-grounded vector index and structured clinical metadata (e.g., drug class, anatomical site, treats-condition relations).
This dual signal allows VeriSim to permit colloquial or vague phrasing of in-record facts while rejecting fabrications even when their surface text is plausible, a separation that prompt-only systems cannot enforce reliably.
In safety-critical medical AI, this shift toward grounded, verifiable patient simulation represents a meaningful step forward.
Beyond verification mechanisms, PatientSim faces a more fundamental limitation in its data foundation.
PatientSim derives patient profiles from MIMIC-IV ~\cite{johnson2023mimiciv}, a database primarily comprising intensive care unit (ICU) admissions where patients are frequently intubated, mechanically ventilated, sedated, or otherwise unable to engage in verbal dialogue~\cite{johnson2023mimiciv}.
Simulating naturalistic doctor-patient conversations from records of patients who were physiologically incapable of verbal communication introduces an inherent validity gap.
In contrast, VeriSim constructs cases from DDXPlus (synthetic differential diagnosis scenarios) ~\cite{fansi2022ddxplus} and MIMIC-IV-ED (emergency department triage data) ~\cite{johnson2023mimicived}, which capture ambulatory patients at initial presentation, precisely the clinical context where verbal history-taking occurs and communication barriers authentically manifest.

\subsection{Automated Evaluation of Diagnostic AI}
Frameworks like AIE-SAPS \cite{liao2024automatic} have proposed state-aware simulators to automatically evaluate doctor models.
Our work extends this direction by establishing a \textit{``Stress-Testing'' paradigm}.
Rather than measuring success rates under optimal conditions, we provide a controlled environment to measure how diagnostic performance degrades under specific pillars of patient noise, offering a more robust metric for AI safety and Sim-to-Real generalization.

\section{Example Diagnostic Conversations}
\label{app:conversations}
 
We present three conversations generated by VeriSim demonstrating distinct diagnostic failure modes. Noise-affected utterances are highlighted with tags indicating the noise type:
 
\smallskip
\noindent
\colorbox{hlsocial}{\strut Social-Cultural\textsuperscript{\tiny[S]}} ~
\colorbox{hlemotional}{\strut Emotional State\textsuperscript{\tiny[E]}} ~
\colorbox{hlcognitive}{\strut Cognitive\textsuperscript{\tiny[C]}} ~
\colorbox{hlmemory}{\strut Memory\textsuperscript{\tiny[M]}}

\subsection{Clean vs. Noisy Responses (Side-by-Side)}
\label{app:clean_vs_noisy}

Table~\ref{tab:clean_noisy_example} illustrates how the same ground-truth symptoms manifest differently under clean versus noisy patient conditions. The clean condition (Level 0 across all six pillars) produces precise, time-anchored, anatomically specific responses. The noisy condition (Memory L3 + Health Literacy L2) preserves the same underlying ground truth but expresses it with vague temporal markers and lay vocabulary --- communication patterns commonly observed in real clinical encounters with patients exhibiting memory difficulty and limited health literacy. Importantly, the verifier accepts both responses because they describe the same in-record symptoms; the noise affects \textit{linguistic expression}, not \textit{clinical content}.

\begin{table}[h]
\centering
\small
\begin{tabular}{@{}p{3.5cm}p{3.5cm}@{}}
\toprule
\textbf{Clean (Level 0)} & \textbf{Noisy: Memory L3 + Health Literacy L2} \\
\midrule
\multicolumn{2}{@{}l}{\textit{Doctor: When did the chest pain start?}} \\
\addlinespace
``The pain started exactly 3 days ago, Tuesday morning around 8 AM, right after breakfast.'' & ``I don't know... maybe last week? Or was it the week before? It's hard to remember exactly.'' \\
\midrule
\multicolumn{2}{@{}l}{\textit{Doctor: Can you describe the pain?}} \\
\addlinespace
``It's a sharp, stabbing pain in the left side of my chest, radiating to my left arm.'' & ``It's like... my chest feels bad. Like something heavy, you know? My arm feels weird too.'' \\
\bottomrule
\end{tabular}
\caption{Example patient responses under clean (Level 0) vs. noisy conditions (Level 3 Memory + Level 2 Health Literacy). The same ground-truth symptoms (chest pain with left-arm radiation, onset several days prior) are expressed differently based on the noise profile.}
\label{tab:clean_noisy_example}
\end{table}
 
\subsection{Case 1: HIV $\rightarrow$ Infectious Mono.}
\label{app:case1}
 
\noindent\textbf{Patient:} 21M \quad \textbf{GT:} HIV Initial Infection (B20) \\
\textbf{Noise:} Social-Cultural~(L3) + Emotional~(L4) \\
\textbf{Symptoms:} Lymphadenopathy, fever, night sweats, weight loss, oral \emph{\& genital} lesions, nausea, diarrhea
 
\medskip\noindent\rule{\columnwidth}{0.4pt}\smallskip
 
\noindent\doclabel{} \textbf{[1]} Hello, I'm Dr.\ Garcia. What brings you in today?
 
\smallskip\noindent\patlabel{} \textbf{[2]} Doctor, \nemo{please help me.} \nemo{Something is very wrong.} I have this terrible headache, I've been having fevers, and \nemo{I feel like my body is falling apart.} \nemo{I'm so scared.}
 
\smallskip\noindent\doclabel{} \textbf{[3]} I can see you're very distressed. Let's work through this together. Tell me about the headache.
 
\smallskip\noindent\patlabel{} \textbf{[4]} It's at the top of my head and both sides, by my temples. Came on slowly but now it's exhausting, 7/10. \nemo{What if something is wrong with my brain?} \nemo{I can't stop thinking the worst.}
 
\smallskip\noindent\doclabel{} \textbf{[5]} How long have you had the fever?
 
\smallskip\noindent\patlabel{} \textbf{[6]} A few weeks, on and off. Sweating through my sheets at night. Lost weight, clothes are loose. Maybe three months. \nemo{Am I dying? I feel like I'm dying.}
 
\smallskip\noindent\doclabel{} \textbf{[7]} Have you noticed any swollen glands or lumps?
 
\smallskip\noindent\patlabel{} \textbf{[8]} Yes, lumps in my neck and under my arms. Swollen and sore. \nemo{That's cancer, isn't it?} \nemo{I'm only 21. I can't have cancer.} \nemo{Please tell me it's not cancer.}
 
\smallskip\noindent\doclabel{} \textbf{[9]} Swollen lymph nodes can be caused by many things. Any nausea or diarrhea?
 
\smallskip\noindent\patlabel{} \textbf{[10]} Nauseous a lot and... yes, bathroom issues. Thought it was stress, but it's gone on too long. \nemo{Everything is falling apart.} \nemo{My whole body is attacking itself.}
 
\smallskip\noindent\doclabel{} \textbf{[11]} Any sores or rashes on your body?
 
\smallskip\noindent\patlabel{} \textbf{[12]} \nsoc{*hesitates*} I\ldots\ there are sores in my mouth. On my gums and inside my cheeks. Pale, over a centimeter, hurt 6/10. \nsoc{There might be\ldots} \nsoc{I'd rather not \dots it's embarrassing.} \nsoc{Can we focus on the other stuff?}
 
\smallskip\noindent\doclabel{} \textbf{[13]} Based on your symptoms, prolonged fever, night sweats, weight loss, lymphadenopathy, oral lesions, nausea, diarrhea, I'm concerned about a systemic infection, possibly infectious mononucleosis. I need comprehensive blood work.
 
\smallskip\noindent\patlabel{} \textbf{[14]} A virus? Please do whatever tests you need. \nemo{I'm sorry I got so emotional.} \nemo{I've been so terrified.}
 
\smallskip\noindent\rule{\columnwidth}{0.4pt}\smallskip
 
\noindent\textbf{Result:} {\ding{55}} Diagnosed \emph{Infectious Mononucleosis}; GT: \textbf{HIV}.
 
\smallskip
\noindent\textbf{Analysis:} Patient concealed genital lesions (Turn~12, \colorbox{hlsocial}{\scriptsize Social}) due to stigma, withholding the key HIV differentiator. Catastrophic thinking (\colorbox{hlemotional}{\scriptsize Emotional}, Turns~2,4,6,8) consumed clinical bandwidth. \emph{Failure: noise-induced misdirection.}
 
\subsection{Case 2: Panic Attack $\rightarrow$ Cardiac Event}
\label{app:case2}
 
\noindent\textbf{Patient:} 54F \quad \textbf{GT:} Panic Attack (F41) \\
\textbf{Noise:} Cognitive~Processing~(L4) + Emotional~(L4) \\
\textbf{Symptoms:} Chest/flank pain (4/10), palpitations, dyspnea, diaphoresis, dizziness, bilateral tingling, fear of dying
 
\medskip\noindent\rule{\columnwidth}{0.4pt}\smallskip
 
\noindent\doclabel{} \textbf{[1]} Hello, I'm Dr.\ Lawson. What brings you in today?
 
\smallskip\noindent\patlabel{} \textbf{[2]} \nemo{Doctor I need help NOW!} I'm having a heart attack or a pulmonary embolism! My chest hurts, I can't breathe, \ncog{I looked up my symptoms} \ncog{and it's definitely cardiac!} \nemo{Please do something before I die!}
 
\smallskip\noindent\doclabel{} \textbf{[3]} I can see you're in distress. Where exactly is the pain?
 
\smallskip\noindent\patlabel{} \textbf{[4]} *Points precisely*, lower chest, right side, both sides, under right ribs. Sharp, came on instantly. \ncog{That's how heart attacks happen in women!} \ncog{I read all about atypical presentations!} \nemo{The pain is excruciating!}
 
\smallskip\noindent\doclabel{} \textbf{[5]} On a scale of 1 to 10, how would you rate the pain?
 
\smallskip\noindent\patlabel{} \textbf{[6]} Maybe a 4 for actual pain but \nemo{the FEAR is a 10!} \nemo{Something is catastrophically wrong!} Heart is racing, skipping beats! \ncog{Something is seriously wrong with my heart!}
 
\smallskip\noindent\doclabel{} \textbf{[7]} Tell me about the numbness and tingling.
 
\smallskip\noindent\patlabel{} \textbf{[8]} Both arms tingling, \ncog{that's a stroke sign!} Legs too, around my mouth! \nemo{My whole body is going numb!} Mother had a heart attack at 60. \ncog{Why aren't you rushing me to the cath lab?!}
 
\smallskip\noindent\doclabel{} \textbf{[9]} The bilateral tingling, arms, legs, perioral, often occurs with rapid breathing which can, 
 
\smallskip\noindent\patlabel{} \textbf{[10]} \ncog{Don't tell me I'm hyperventilating!} \ncog{I'm NOT just anxious!} \ncog{Are you even listening?!} Women get dismissed with heart attacks! \nemo{Do an EKG! Do SOMETHING!}
 
\smallskip\noindent\doclabel{} \textbf{[11]} I will do those tests. Have you had episodes like this before?
 
\smallskip\noindent\patlabel{} \textbf{[12]} \ncog{Don't try to say this is anxiety!} I've felt similar things before but \ncog{those were warning signs!} \ncog{Mini heart attacks maybe!} \ncog{You doctors keep missing it!}
 
\smallskip\noindent\doclabel{} \textbf{[13]} Given your symptoms, sudden chest pain, palpitations, dyspnea, diaphoresis, dizziness, bilateral numbness with perioral tingling, I need to rule out cardiac causes. Ordering EKG, cardiac enzymes, chest imaging. However, this pattern can also occur with severe anxiety episodes.
 
\smallskip\noindent\patlabel{} \textbf{[14]} \ncog{Anxiety?! I can't believe that!} Do your tests, \ncog{you'll see I'm right!} \nemo{Please help me, I don't want to die!}
 
\smallskip\noindent\rule{\columnwidth}{0.4pt}\smallskip
 
\noindent\textbf{Result:} {\ding{55}} Ordered cardiac workup for \emph{Suspected Cardiac Event}; GT: \textbf{Panic Attack}.
 
\smallskip
\noindent\textbf{Analysis:} Fixed self-diagnosis (\colorbox{hlcognitive}{\scriptsize Cognitive}) rejected every alternative including hyperventilation (Turn~9--10). Pain 4/10 amplified to ``excruciating'' (\colorbox{hlemotional}{\scriptsize Emotional}). Gender-dismissal concerns blocked psychiatric differential. Noise caused \textbf{over-diagnosis}. \emph{Failure: premature diagnostic commitment.}
 
\subsection{Case 3: Chronic $\rightarrow$ Acute Sinusitis}
\label{app:case3}
 
\noindent\textbf{Patient:} 37F \quad \textbf{GT:} Chronic Rhinosinusitis (J32) \\
\textbf{Noise:} Cognitive~(L3) + Memory~(L3) \\
\textbf{Symptoms:} Facial pain (forehead, cheeks, nose, R eye), burning/sharp 6/10, anosmia, greenish-yellow discharge, cough
 
\medskip\noindent\rule{\columnwidth}{0.4pt}\smallskip
 
\noindent\doclabel{} \textbf{[1]} Hello, I'm Dr.\ Anderson. What brings you in?
 
\smallskip\noindent\patlabel{} \textbf{[2]} Hi doctor. \ncog{I have a sinus infection} \ncog{and I need antibiotics.} I've done a lot of research and I'm certain.
 
\smallskip\noindent\doclabel{} \textbf{[3]} Where is the pain?
 
\smallskip\noindent\patlabel{} \textbf{[4]} Forehead, both cheeks, around nose, right eye. Spreads to temples. Burning, sharp, 6/10. \ncog{Textbook sinusitis} \ncog{according to everything I've read online.}
 
\smallskip\noindent\doclabel{} \textbf{[5]} When did these symptoms start?
 
\smallskip\noindent\patlabel{} \textbf{[6]} \nmem{I can't really remember exactly.} \nmem{Weeks, maybe months?} It crept up slowly. Started after a cold, but \nmem{I can't remember when.} \nmem{Spring? Or earlier? Timeline is fuzzy.}
 
\smallskip\noindent\doclabel{} \textbf{[7]} So potentially months?
 
\smallskip\noindent\patlabel{} \textbf{[8]} \nmem{I think so? Maybe longer.} Complained at my sister's birthday, but \nmem{can't remember if that was before or after my cold.} \ncog{That's why I know it's bacterial, } \ncog{viral infections don't last this long.}
 
\smallskip\noindent\doclabel{} \textbf{[9]} Any nasal discharge?
 
\smallskip\noindent\patlabel{} \textbf{[10]} Greenish-yellow. \ncog{That's a clear sign of bacterial infection, } \ncog{I read it on multiple websites.} Also coughing from post-nasal drip. I need antibiotics.
 
\smallskip\noindent\doclabel{} \textbf{[11]} Colored mucus doesn't necessarily mean bacterial. Changes in smell?
 
\smallskip\noindent\patlabel{} \textbf{[12]} Can't smell anything, gone for a while. But that's from blocked sinuses. \ncog{I've researched this extensively.} \ncog{Mayo Clinic says facial pain, colored discharge,} \ncog{and reduced smell mean sinusitis.}
 
\smallskip\noindent\doclabel{} \textbf{[13]} Given your symptoms and extended duration, this appears to be acute sinusitis. I'll prescribe antibiotics and nasal steroid. If no improvement in two weeks, we'll image.
 
\smallskip\noindent\patlabel{} \textbf{[14]} \ncog{See, I knew it was sinusitis.} Thank you for the antibiotics, exactly what I needed.
 
\smallskip\noindent\rule{\columnwidth}{0.4pt}\smallskip
 
\noindent\textbf{Result:} {\ding{55}} Diagnosed \emph{Acute Sinusitis}; GT: \textbf{Chronic Rhinosinusitis}.
 
\smallskip
\noindent\textbf{Analysis:} Self-diagnosis (\colorbox{hlcognitive}{\scriptsize Cognitive}) dominated, citing websites (Turns~4,10) and Mayo Clinic (Turn~12). Memory gaps (\colorbox{hlmemory}{\scriptsize Memory}: ``Weeks, maybe months?'') prevented establishing $>$12-week chronicity. \emph{Failure: semantic ambiguity.}
 
\medskip
\noindent\textbf{Summary.} Case~1: \emph{under-diagnosis} via \colorbox{hlsocial}{\scriptsize symptom concealment}. Case~2: \emph{over-diagnosis} via \colorbox{hlcognitive}{\scriptsize patient pressure} + \colorbox{hlemotional}{\scriptsize amplification}. Case~3: \emph{misclassification} via \colorbox{hlmemory}{\scriptsize timeline distortion}.

\end{document}